\documentclass[sigconf]{acmart}
\usepackage{subcaption}
\AtBeginDocument{%
  }

\copyrightyear{2025}
\acmYear{2025}
\setcopyright{cc}
\setcctype{by}
\acmConference[FAccT '25]{The 2025 ACM Conference on Fairness,
Accountability, and Transparency}{June 23--26, 2025}{Athens, Greece}
\acmBooktitle{The 2025 ACM Conference on Fairness, Accountability, and
Transparency (FAccT '25), June 23--26, 2025, Athens,
Greece}\acmDOI{10.1145/3715275.3732029}
\acmISBN{979-8-4007-1482-5/2025/06}

\begin{document}

\title{Artifacts of Idiosyncracy in Global Street View Data.}

\author{Tim Alpherts}
\email{t.o.l.alpherts@uva.nl}
\affiliation{%
  \institution{University of Amsterdam}
  \city{Amsterdam}
  \country{The Netherlands}
}
\author{Sennay Ghebreab}
\email{s.ghebreab@uva.nl}
\affiliation{%
  \institution{University of Amsterdam}
  \city{Amsterdam}
  \country{The Netherlands}
}
\author{Nanne Van Noord}
\email{n.j.e.vannoord@uva.nl}
\affiliation{%
  \institution{University of Amsterdam}
  \city{Amsterdam}
  \country{The Netherlands}
}

\renewcommand{\shortauthors}{Alpherts et al.}

\begin{abstract}
Street view data is increasingly being used in computer vision applications in recent years. Machine learning datasets are collected for these applications using simple sampling techniques. These datasets are assumed to be a systematic representation of cities, especially when densely sampled. Prior works however, show that there are clear gaps in coverage, with certain cities or regions being covered poorly or not at all. Here we demonstrate that a cities' idiosyncracies, such as city layout, may lead to biases in street view data for 28 cities across the globe, even when they are densely covered. We quantitatively uncover biases in the distribution of coverage of street view data and propose a method for evaluation of such distributions to get better insight in idiosyncracies in a cities' coverage. In addition, we perform a case study of Amsterdam with semi-structured interviews, showing how idiosyncracies of the collection process impact representation of cities and regions and allowing us to address biases at their source.
\end{abstract}

\begin{teaserfigure}
\centering
  \includegraphics[width=.7\textwidth]{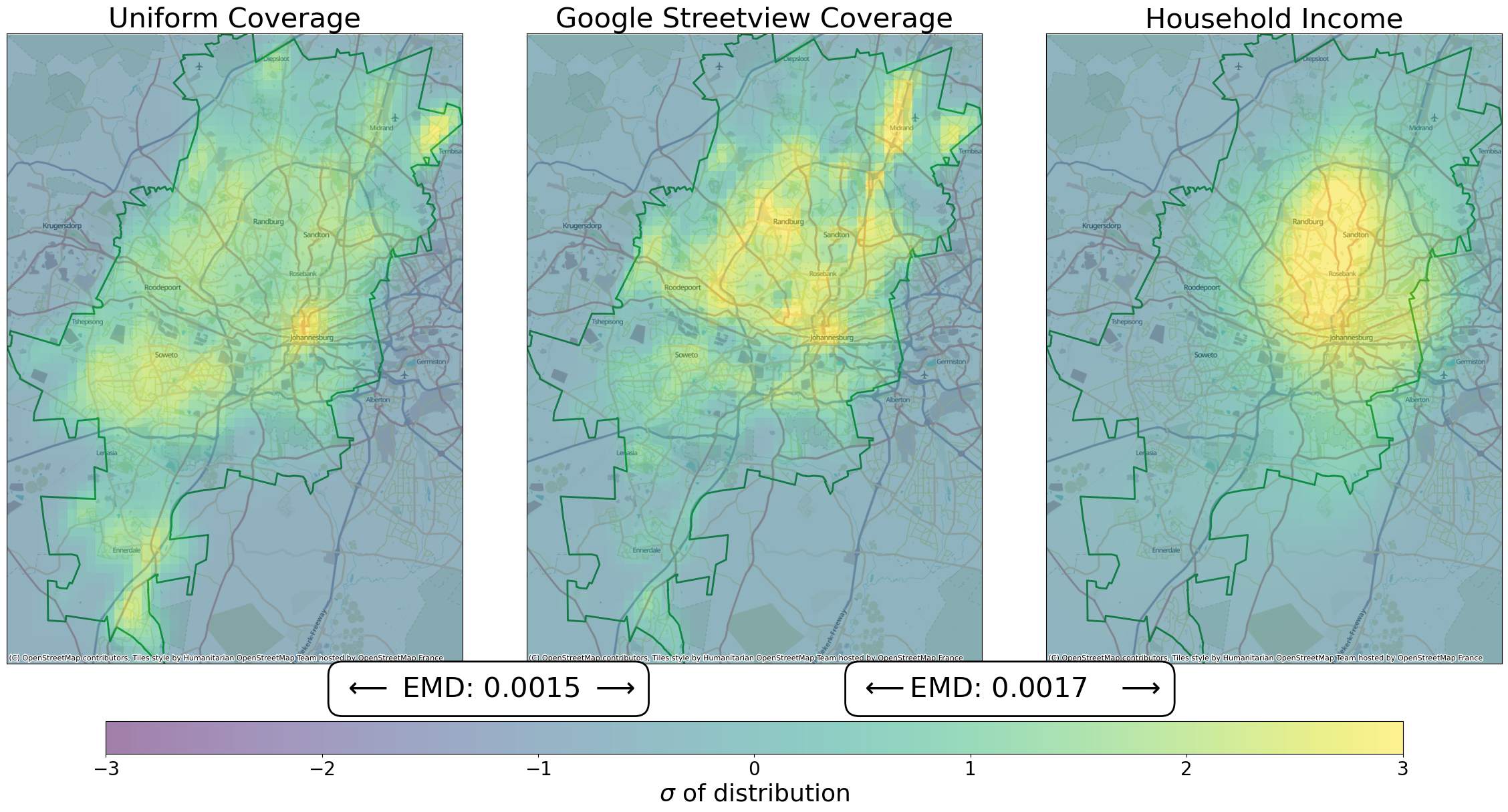}
  \Description{Spatial distribution of Socio-Economic Statistics alongside spatial distribution of Google Street View images. Previous methods mark Johannesburg as having near perfect coverage. Idiosyncracies in driving patterns, data collection, or city dynamics can unknowingly create artifacts of inequality in street view datasets. }
  \label{fig:teaser}
  \caption{Analysis of the available Google Street View in Johannesburg (middle) compared against a uniform prior over the road network (left) and a prior based on the mean household income over 2016-2021 \cite{gcro2016qol4, gcro2022qol5, gcro2021qol6, hamann2024segregation} (right). From the comparable Earth Mover's distance (EMD) we can observe that the distribution of street view imagery is explained equally well by either prior. This highlights how the idiosyncrasies of a city propagate to the available street view data, and to AI applications trained on such data.}
\end{teaserfigure}

\maketitle

\section{Introduction}
Since the introduction of Google Street View the AI community has looked to leverage street view images to train models. The promise of using computer vision to automate processes in the city or to support urban planners, in combination with the abundance of street view imagery has led to new fields of research and a torrent of urban vision papers. Computer vision models have been developed for tasks such as Visual Place Recognition \cite{Arandjelovi, iovpr, Camara2019, Cao2015, Torii2013, Gomez-Ojeda2015}, Visual Urban Analytics \cite{alphertsPerceptiveVisualUrban2024, Ordonez2014, Suel2019, Naik2014, Arietta2014, Seresinhe2017, Dubey2016, Law2019, Gebru2017}, Urban Scene Change Detection \cite{Sakurada2018, huangCityPulseFineGrainedAssessment2024, Alcantarilla2016}, or for monitoring objects in the cityscape such as potholes \cite{Ma2022}, waste \cite{Sukel2020}, or trees \cite{Seiferling2017}. Moreover, Computer Vision models have been used to explore theories in Sociology regarding neighbourhood appearance and socio-economic indicators in the form of prediction of labels such as mean income \cite{Suel2019}, scenicness \cite{Seresinhe2017}, and theft \cite{Arietta2014} or crime rates \cite{Fu2018}. 
The datasets these models are trained on are constructed from providers such as Google Street View \cite{GoogleLandmarksV2, Suel2019} or Mapillary \cite{vistas}. A crucial assumption underlying these studies is that
collection of street view images follows a mechanistic approach and as a consequence the constructed datasets use sampling at fixed spatial intervals \cite{Muller2022, Groenen2022} to "get a contiguous visual representation of the city" \cite{Muller2022}. In this paper we question this assumption and show that that there are clear biases present within street view data.

Previous studies into street view data have evaluated the representation of cities by street view providers with spatial coverage: they evaluated whether a road or intersection is covered within the street view imagery. Some cities have street view imagery on all streets while in other cities capturing is restricted to highways or major arteries \cite{quinnEverySingleStreet2019}. This binary coverage definition has led to the discovery of coverage differences between urban and rural areas \cite{machicao2022}, partially due to the different infrastructure \cite{kimExaminationSpatialCoverage2023}. The implications of these findings are that we may observe biases in tasks that are applied to both rural areas and urban areas. 

While prior studies discuss the issue of coverage biases they are constrained to noting that certain cities are partially covered. Large cities in which there is evidence of systematic coverage, such as Johannesburg \cite{quinnEverySingleStreet2019}, are found to have the highest coverage level and are as such considered suitable for computer vision methods. However, checking for whether a road has been covered disregards the potential of certain roads being covered more \textit{often} than others. Parts of the city that are more easily accessible, or that lie near entrance points to other neighbourhoods might be covered more often due to the nature of the collection process. Previous works focused solely on spatial coverage, while not considering differences in the spatial \textit{distribution} of coverage. When constructing machine learning datasets, depending on how the sampling is done (e.g., multiple images per location, or only the most recent image), the differences in coverage rates across neighbourhoods will be reflected in the resulting dataset. Such differences in coverage density may be reflective of underlying systemic biases which would then propagate to downstream applications trained upon this data.

To investigate additional sources of bias in streetview imagery we focus, not on spatial coverage, but on the spatial \textit{distribution} of coverage. We evaluate the coverage of street view providers with respect to what uniform road coverage would look like. We show how the distribution of street view data with respect to an uniform distribution affects a variety of global cities in different ways. We compare distance metrics for measuring the difference in distributions and we show that good coverage does not necessarily mean equal distribution of data.
Our contributions are as follows:

\begin{itemize}
    \item We quantitatively uncover biases in the distribution of coverage of street view data. 
    \item We propose a method for evaluating the distribution of street view data, which gives insight into the idiosyncrasies of a cities' coverage.
    \item We show in a case study of Amsterdam through (n=6) semi-structured interviews that idiosyncracies on all levels of the collection process can lead to coverage biases. 
\end{itemize}

\section{Related Work}
\subsection{Measuring Street View Coverage}

The earliest works on measuring coverage of street view providers are more than a decade old, yet, these works have been sporadic and primarily from the Geographic Information System (GIS) community \cite{curtisUsingGoogleStreet2013,rzotkiewiczSystematicReviewUse2018,umarPotentialGoogleStreet2023}. As such, these studies have mainly focused on the impact of the available data for manual auditing purposes. The earliest example of this mentioned temporal instability as a problem for auditing purposes \cite{curtisUsingGoogleStreet2013}. As different parts of the city are driven at different times, continuity around intersections could vary and images could skip ahead or back in time. Subsequent studies have focused on more specific questions.  
\cite{rzotkiewiczSystematicReviewUse2018} evaluated the feasibility of using Google Street View for health research. As ``image data is updated by Google at a frequency that is dependent on population density and weather conditions'' they conclude that rural areas may have less coverage than urban areas.
\cite{umarPotentialGoogleStreet2023} evaluated the coverage of GSV on the African continent for waste collection purposes, focusing on the availability of images for waste collection sites. A people-based approach is taken by \cite{kimExaminationSpatialCoverage2023}, looking at coverage from the perspective of commute trajectories in 45 small- and medium-sized cities in the US. Differences in coverage patterns between cities as a whole have been previously looked at by two studies. Image availability was evaluated for 371 Latin American cities by \cite{fryAssessingGoogleStreet2020}. \cite{quinnEverySingleStreet2019} performed a similar study but at a global scale including Mapillary and Open Street Map alongside Google Street View, creating a rating scale to classify how well a city was covered. However, for both these studies the coverage was evaluated using an available/not available classification. Instead, in our work we are interested in the \textit{distribution} of coverage in cities. 

The first study to look at coverage from an angle of distribution is \cite{gsvbronx} in a 2000 image study to asses potential bias in change auditing using Google Street View in images over the years. This method uses manual analysis through annotation by multiple auditors, which is is labour-intensive and not scalable. Furthermore this differs from our paper as no distributions were calculated, nor were the observed trends identified in more than two neighbourhoods. Yet, the findings in this work motivate the urgency of our research into creating a scalable methodology. 

\subsection{Street View Datasets for AI}
Within AI research an abundance of datasets have been constructed from street view data using data from different providers. Datasets are primarily sourced from Google Street View, such as GSV-Cities \cite{Ali_bey_2022}, SF-XL \cite{berton2022rethinkingvisualgeolocalizationlargescale}, SVOX \cite{Moreno_Berton_2021}, or the Pittsburgh250k \cite{Torii2013} which are used for Visual Place Recognition. Approaches for Perceptive Visual Urban Analytics include Place Pulse 1.0 \cite{Naik2014}, Place Pulse 2.0 \cite{Dubey2016}, or the London dataset used by \cite{Suel2019}. 
Mapillary datasets are used in multiple studies including Mapillary Vistas \cite{Neuhold_2017_ICCV}, SLS \cite{Warburg}, Traffic Sign Dataset \cite{ertler2020mapillarytrafficsigndataset}, Road Surface Global Dataset \cite{randhawa2024pavedunpaveddeeplearning}. 

A number of papers introduce datasets from other sources such as Amos \cite{jacobs09webcamgis}, Urban Mosaic \cite{Miranda2020}, Flickr \cite{iovpr}, and the Amsterdam Panorama Database \cite{Groenen2022}; or from more dense data streams such as dashcam data \cite{Franchi_2023, che2019d2citylargescaledashcamvideo, moura2025nexardashcamcollisionprediction}. For our purposes we choose to restrict our scope to three static databases: Google Street View, Mapillary, and a single private database. This to cover the two main providers of street view datasets, and a case study into the dynamics of a private provider. Furthermore, while OpenStreetCam has previously been evaluated as source of images in previous studies \cite{quinnEverySingleStreet2019} we exclude this as we were unable to find relevant AI papers using this imagery. 

For constructing street view image datasets three main sampling methods have been utilized: using all images \cite{Torii}, uniform sampling to ensure geographical spread \cite{Muller2022}, and taking all images for a set amount of locations \cite{Arandjelovi, huangCityPulseFineGrainedAssessment2024, alphertsPerceptiveVisualUrban2024}. All three methods interact with the distribution of street view data in a different manner. Uniform sampling suffers from a recency bias, where neighbourhoods that haven't been driven in a long time have images that are outdated as opposed to neighbourhoods driven more recently. Taking a cluster of images for a set amount of location is used in Urban Scene Change Detection or Visual Place Recognition. To learn differences between two images per location, these clusters are turned into image pairs for training purposes, resulting in $\binom{n}{2}$ image pairs for a cluster of size n. This quickly scales the number of datapoints when clusters increase in size, which may lead to biases towards areas for which there is more imagery, as it is easier to construct clusters here. Finally, when taking all available images the data distribution stays the same thus mirroring any existing biases.

\section{Method}

\begin{figure}
    \centering
    \includegraphics[width=.95\linewidth]{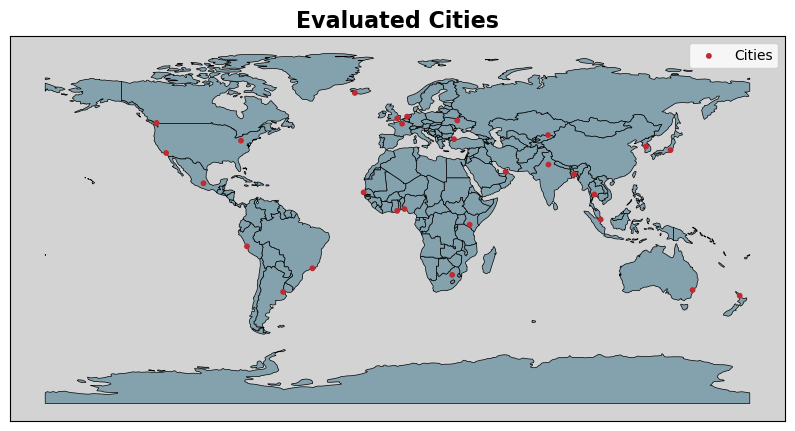}
    \caption{Overview of cities where the coverage distribution was evaluated. Cities were picked to ensure a good geographical spread.}
    \label{fig:worldcities}
\end{figure}

Our goal is to evaluate the coverage distribution of street view providers for cities across the globe. For this we will evaluate both Google Street View and Mapillary, as both have been widely used for AI research and operate globally. Our analysis focuses on 28 cities spread across continents, as shown in Figure~\ref{fig:worldcities}. Cities were selected with the aim to have variety in geographical location, continent, size of the metropolitan area, and use within AI datasets. Cities in North Africa as well as China were excluded due to a lack of coverage. We collected the street view data for each city in a systematic manner.

The foundation of our method is to compare the actual coverage distribution of a city based on street view data to a prior distribution, by measuring the deviation from this prior we are able to uncover biases. To establish this prior we consider a uniform distribution across the road network of the city, that is, each road is expected to be covered at the same frequency. We consider a uniform prior as the collection of street view imagery is often assumed to be a mechanistic process with sampling at fixed spatial intervals  \cite{Muller2022, Groenen2022}, in which case the resulting data would be uniformly distributed. However, if there are deviations from this uniform coverage then these give insight into if, and where, there are biases in the street view data.

Our expectation is that the deviations found for a city are influenced by idiosyncrasies of that city and the collection process. The resulting biases may in that respect be of varying nature, for instance, we may observe that major arteries are imaged more frequently. However, differences in imaging frequency may also relate to underlying systematic biases. To give insight in such potential biases our method consists of three stages (1) street view data collection, (2) density estimation, (3) density comparison, each of these stages are explained in the following.

\begin{figure*}
    \centering
    \includegraphics[width=.9\linewidth]{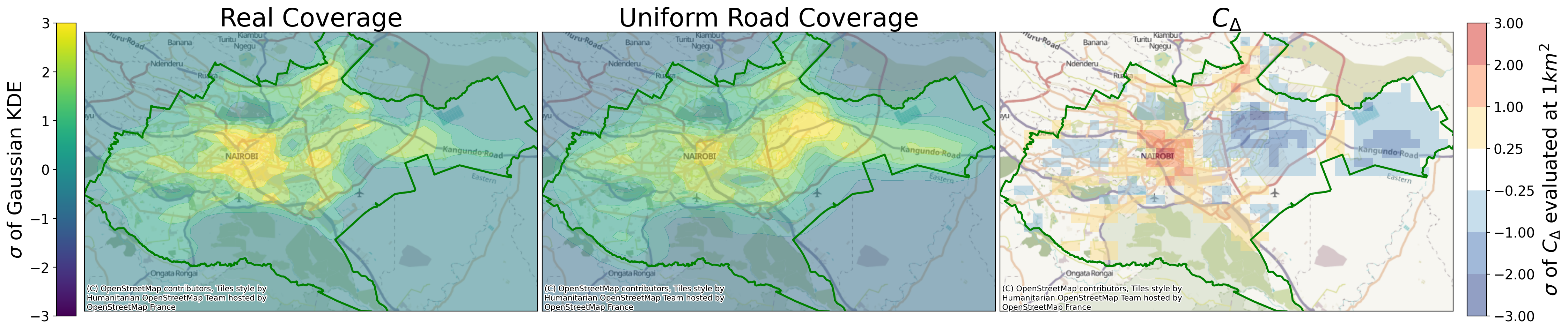}
    \caption{From left to right for Nairobi: Distribution of retrieved metadata in Google Street View, uniform coverage based on available drivable streets in OpenStreetMap, the difference of these two distributions $C_\Delta$ indicating the parts of the city that are oversampled or undersampled.}
    \label{fig:realvsgolden}
\end{figure*}

\subsection{Street View data collection}
In order to retrieve the data for each city we performed the following steps: 
\begin{enumerate}
    \item We obtained the cities' administrative boundary polygons from OpenStreetMap \cite{OpenStreetMap}.
    \item We then retrieve the \textit{metadata} of all images available per city. This process is slightly different for each provider: \begin{itemize}
    \item For Google Street View we construct a spatial grid of points at 20m intervals \cite{Muller2022} across both latitude and longitude and overlay the spatial grid with the polygon. API requests are then made at every point for the closest images within 100m to ensure we retrieve all metadata \cite{fryAssessingGoogleStreet2020}.
    \item For Mapillary we divide the polygon into squares of $400m^2$ meters and retrieve all the metadata within it. 
    \item The Amsterdam municipal database only allows for retrieving images within a radius of a point. We therefore construct a grid of points at 280 meters from each other and request all images within a range of 200 meters. This creates a grid of circles with radius of 200 meters that have 120 meters of overlap on the horizontal and vertical and 5 meters of overlap on the diagonal.
    \end{itemize}
    \item Due to the collection process for each provider this results in duplicate data points; identical panorama IDs that have been retrieved in separate requests due to overlapping boundaries. As such, for each city, provider, duplicates are filtered by these unique IDs. 
    \item Finally, we structure the metadata into a spatial dataset within the city boundary polygon.
\end{enumerate}
All further analysis is performed on the spatial dataset containing the metadata per city.


\subsection{Density Estimation}

If coverage of cities indeed follows a mechanistic process of driving all streets within the cities boundaries we would expect all roads to be driven equally. As such we first create a density map of what this mechanistic coverage would look like. We refer to this as \textit{Uniform Road Coverage}. Note that the density of a location is a function of how often a location has been driven, thus accounting for the temporal granularity. 
To achieve this density map we obtain the streets in each polygon from OpenStreetMap. This graph network is then replaced by evenly spaced points at 20m intervals after which we apply a Gaussian kernel density estimation to calculate a probability density function (PDF) of the data. This PDF is shown on the left in Figure~\ref{fig:realvsgolden}. Subsequently, we calculate the PDF of the data we collected through the street view providers, an example for this is shown in the middle in Figure~\ref{fig:realvsgolden}.

Based on this we have two density estimations for each city: $C_{\textit{Uniform}}$, an estimation of uniform road coverage through the streets from OpenStreetMap and $C_{\textit{Real}}$ an estimation of actual coverage through our pulled panoramas. The difference between $C_{\textit{Real}} - C_{\textit{Uniform}}$ then results in $C_{\Delta}$, a map of the oversampling and undersampling in coverage. A visualization of $C_{\Delta}$ can be seen on the right in Figure~\ref{fig:realvsgolden}. In all visualizations in this paper $C_{\Delta}$ is evaluated on a grid of $1km^2$ squares.  


\subsection{Comparing Distributions} 

\begin{figure*}
    \centering
    \includegraphics[width=.61\linewidth]{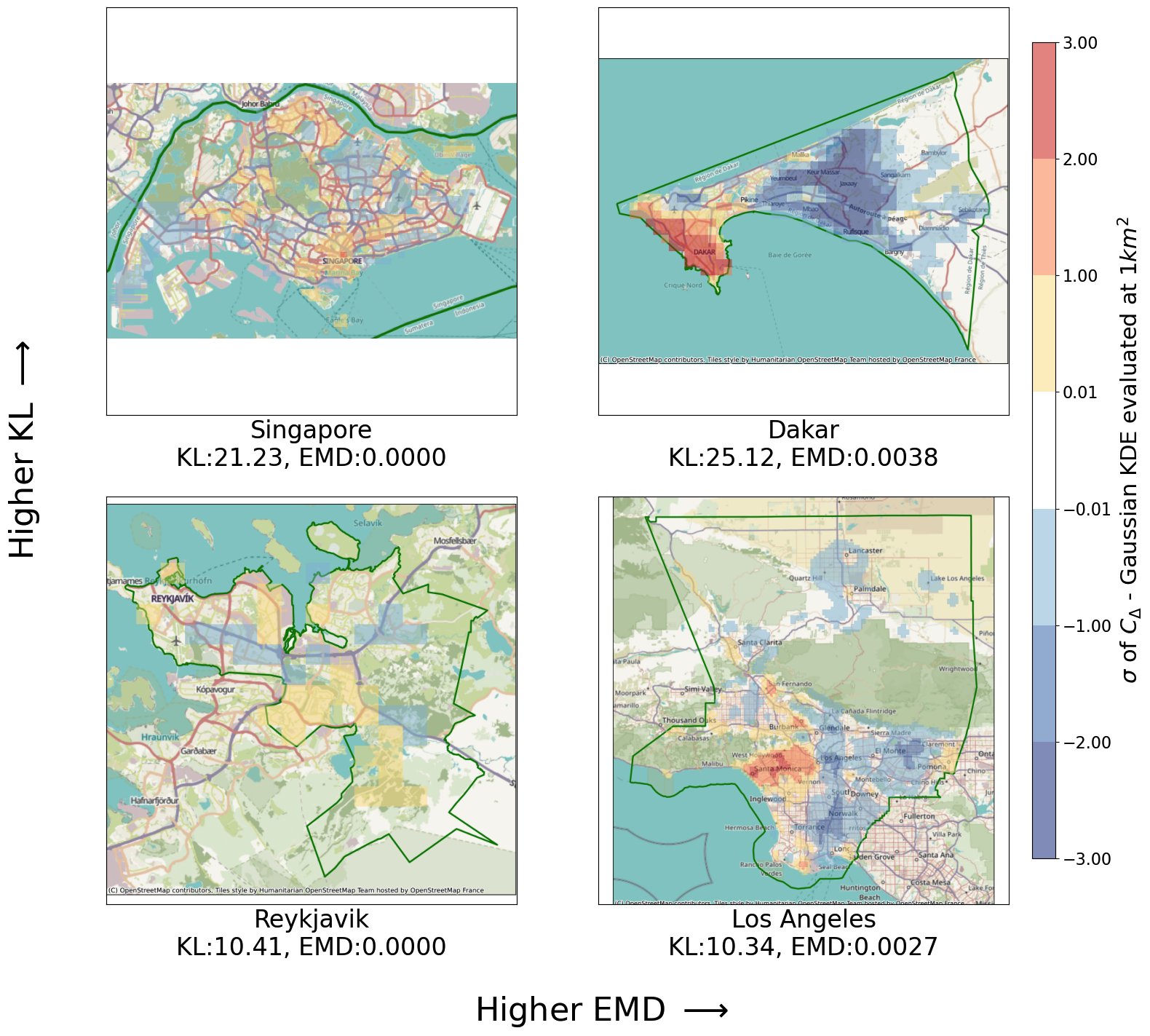}
    \caption{Visualisation of how KL-Divergence and Earth Mover's Distance capture the differences between street view coverage distributions. Top left: For Singapore (GSV) the distribution is not uniform, but the over and undersampled areas are diffusely distributed throughout the city. Top right: For Dakar (Mapillary) coverage is almost only available in the western part of the city. The EMD is high because the centers of mass are distant. Bottom right: The coverage differences in Los Angeles (GSV) are contained to neighbourhoods, but the distrbutions have high overlap as the entire city has coverage. Bottom left: Reykjavik (GSV) has close to uniform coverage and the over and undersampled areas are diffusely distributed throughout the city.} 
    \label{fig:KL_EMD_Methodplot}
\end{figure*}

To measure the difference in distribution between $C_{\textit{Uniform}}$ and $C_{\textit{Real}}$ we use two distance metrics. By quantifying the difference numerically we can provide a ranking for the coverage distribution of different cities. As we are dealing with multivariate distributions we use the following two distance metrics:

\begin{itemize}
    \item The KL-Divergence, as defined by a k-nearest-neighbour density estimation $\hat{D}_k$ \cite{perez-kl2008} . For $n$ i.i.d samples from $p(x)$, $\mathcal{X}=\{x_i\}_{i=1}^n$, and m i.i.d samples from q(x), $\mathcal{X'}=\{x_j'\}_{j=1}^m$: $$\widehat{D}_k(P||Q) = \frac{1}{n}\sum_{i=1}^{n}\text{log}\frac{\widehat{p}_k(x_i)}{\widehat{q}_k(x_i)} = \frac{d}{n}\sum_{i=1}^{n}\text{log}\frac{s_k(x_i)}{r_k(x_i)} + log\frac{m}{n-1}$$ \\
    where
    $$\widehat{p}_k(x_i) = \frac{k}{n-1}\frac{\Gamma(d/2+1)}{\pi^{d/2}r_k(x_i)^d}$$\\
    $$\widehat{q}_k(x_i) = \frac{k}{m}\frac{\Gamma(d/2+1)}{\pi^{d/2}s_k(x_i)^d}$$\\
    Here, $r_k(x_i)$ and $s_k(x_i)$ are the Euclidean distances to the $k^{th}$ nearest-neighbour of $x_i$ in $\mathcal{X}\setminus x_i$ and $\mathcal{X'}$, and $\pi^{d/2}/\Gamma(d/2 + 1)$ is the volume of the unit-ball in $\mathbb{R}^d$. 

    For more details and a proof we refer to \cite{perez-kl2008}. The intuition behind using the KL-Divergence is that it measures the amount of overlap between two distributions, in our case the distribution of available data and the distribution of Uniform Road Coverage. In practice this means that the KL-Divergence is sensitive to roads being skipped in the coverage process. 

    \item Earth Mover's Distance (EMD) \cite{wasserstein}. The EMD can be understood as a transport optimization problem; How much dirt needs to be moved from one pile of the distribution to the other. We use it in addition to the KL-Divergence because it is able to quantify the distance between centers of mass where the KL-Divergence is invariate. 
    
    For $P$ and $Q$ with samples $X_1,...,X_n$ and $Y_1,...,Y_n$ respectively it is defined as: $$W_p(P,Q) = \inf_\pi\left( \frac{1}{n}\sum_{i=1}^{n}||X_{(i)} - Y_{\pi(i)}|| \right)^{1/p}$$ where the lower bound is calculated over all permutations of $\pi$ of $n$ elements. 
    
    As this is solved with Linear Programming with $O(n^3)$ and we are working with millions of datapoints, we use the Debiased Sinkhorn Divergence \cite{feydy2018interpolatingoptimaltransportmmd} to approximate the EMD. An illustration of the differences between the KL-Divergence and EMD are shown in Fig~\ref{fig:KL_EMD_Methodplot}: We see that the EMD is more sensitive to large portions of mass moving further away from each other, while The KL is more sensitive to mass covering disjointed areas. In practice this means: A low KL indicates all roads have been covered, while a high KL indicates there are gaps in coverage. A low EMD indicates all parts of the city are being covered equally, while a high EMD indicates most of the coverage occurs in certain neighbourhoods.
    
\end{itemize}

\section{Coverage Distribution Analysis}

We present the results of our analysis in three stages. First we evaluate the coverage distributions using the EMD and KL and provide a new ranking for street view databases based on these scores.
Secondly, we perform an analysis to evaluate whether the coverage distribution is a relevant metric. We evaluate the correlation between both EMD and KL with the coverage percentages. A strong correlation, (>.7) would indicate that evaluating coverage by percentage is enough. Less would indicate that evaluating the coverage distribution of the city through metrics such as EMD and KL is necessary to give further insight into the coverage. Furthermore, OpenStreetMap makes a distinction between driveable and all publicly accessible roads. In some cities, certain neighbourhoods do not have roads accessible by cars. Therefore street view providers sometimes make the effort to collect data by bike or backpack. To evaluate whether this is done in an equal manner we repeat the experiments for both type of roads. 
Finally, we perform a qualitative analysis of our method by looking at what the $C_{\Delta}$ maps can tell us about the distribution of street view coverage throughout the city.
\begin{figure*}
    \includegraphics[width=.87\linewidth]{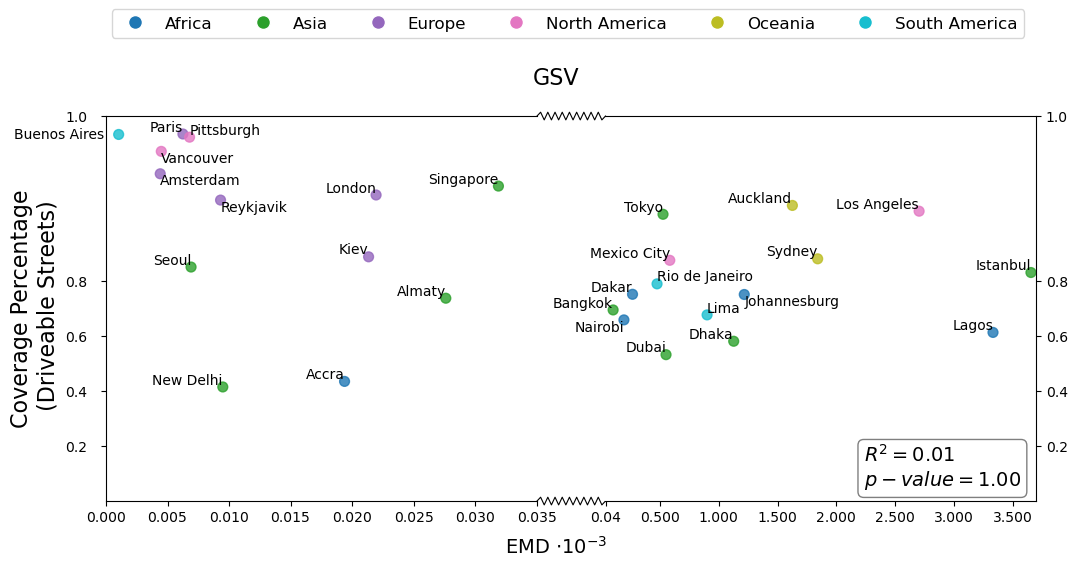}
    \caption{Coverage percentages plotted against the Earth Mover's Distance for Google Street View. Note that the Y and X axis are respectively extended and compressed for readability.}
    \label{fig:worldcities_covcorplot}
\end{figure*}

\begin{table}[hb!]
    \centering
    \scriptsize
    \begin{tabular}{llrrrr}
    \toprule
    City & Provider & EMD, $10^{-3}$ & KL & EMD-Rank & KL-Rank \\
    \midrule
Kiev & GSV & .021 & 7.55 & 13 & 1 \\
Almaty & GSV & .028 & 9.22 & 15 & 2 \\
Kiev & MLY & .146 & 9.59 & 21 & 3 \\
Los Angeles & GSV & 2.71 & 10.3 & 46 & 4 \\
Auckland & GSV & 1.63 & 10.3 & 43 & 5 \\
Reykjavik & GSV & .009 & 10.4 & 8 & 6 \\
Sydney & GSV & 1.84 & 10.5 & 44 & 7 \\
Pittsburgh & GSV & .007 & 12.1 & 6 & 8 \\
Istanbul & GSV & 3.66 & 12.2 & 50 & 9 \\
Lagos & GSV & 3.33 & 13.2 & 49 & 10 \\
Nairobi & GSV & .192 & 13.3 & 23 & 11 \\
    \multicolumn{6}{c}{\vdots} \\
Mexico City & MLY & 1.47 & 19.4 & 41 & 47 \\
Dakar & GSV & .265 & 20.4 & 25 & 48 \\
Paris & MLY & .058 & 20.8 & 17 & 49 \\
Buenos Aires & MLY & .011 & 21.2 & 11 & 50 \\
Lima & MLY & 1.60 & 21.3 & 42 & 51 \\
Seoul & MLY & .522 & 21.9 & 29 & 52 \\
Tokyo & MLY & .788 & 21.9 & 33 & 53 \\
Singapore & GSV & .032 & 23.5 & 16 & 54 \\
Dakar & MLY & 3.81 & 25.1 & 51 & 55 \\
Singapore & MLY & .255 & 25.7 & 24 & 56 \\
    \bottomrule
    \end{tabular}
    \caption{Results of the evaluation of the distance between Uniform Road Coverage and Real coverage for our selected cities. Scores are ranked individually. The table is sorted based on KL Rank.}
    \label{tab:EMD-KL-Table}
\end{table}

\subsection{Coverage Distributions}
We pulled the street view data for all 28 cities, created the data distribution of Uniform Road Coverage through Open Street Map, and calculated the EMD and KL based on these two distributions for each city respectively.
The results for the distance metrics of coverage distributions for all cities are shown in Table~\ref{tab:EMD-KL-Table}. The coverage distribution per city is ranked based on KL ranking, as this measures how well the shape of the distribution of available data matches the shape of Uniform Road Coverage for that specific city. The full table is shown in the Appendix in Table~\ref{tab:FullRank-Table}. The full table for all publicly accessibly streets is shown in the Appendix in Table~\ref{tab:FullRank-Table_public}.

A number of observations can be made in regards to these results. First off, we note that cities such as Los Angeles, Auckland, have low KL but high EMD. This indicates that over,- and undersampling of data occurs in specific neighbourhoods, rather than spread across different areas in the city. As such, naive uniform sampling from Los Angeles can incur strong biases towards oversampled neighbourhoods for which more data is available. Attention to this is crucial when constructing datasets to prevent a skewed distribution.

Secondly, we observe that the coverage distribution of available data retrieved through Mapillary in general scores much lower than for the same cities retrieved through Google Street View. This indicates that using the data from Google Street View may be a better foundation for further research, yet we still observe strong differences between cities, which indicates that careful consideration remains necessary regardless of the source. Additionally, significance scores for all distributions are calculated using a MANOVA and are shown in the Appendix in Table~\ref{tab:sig-gsv-drive}-\ref{tab:sig-mly-public}.

\begin{figure*}
    \centering
    \includegraphics[width=.82\linewidth]{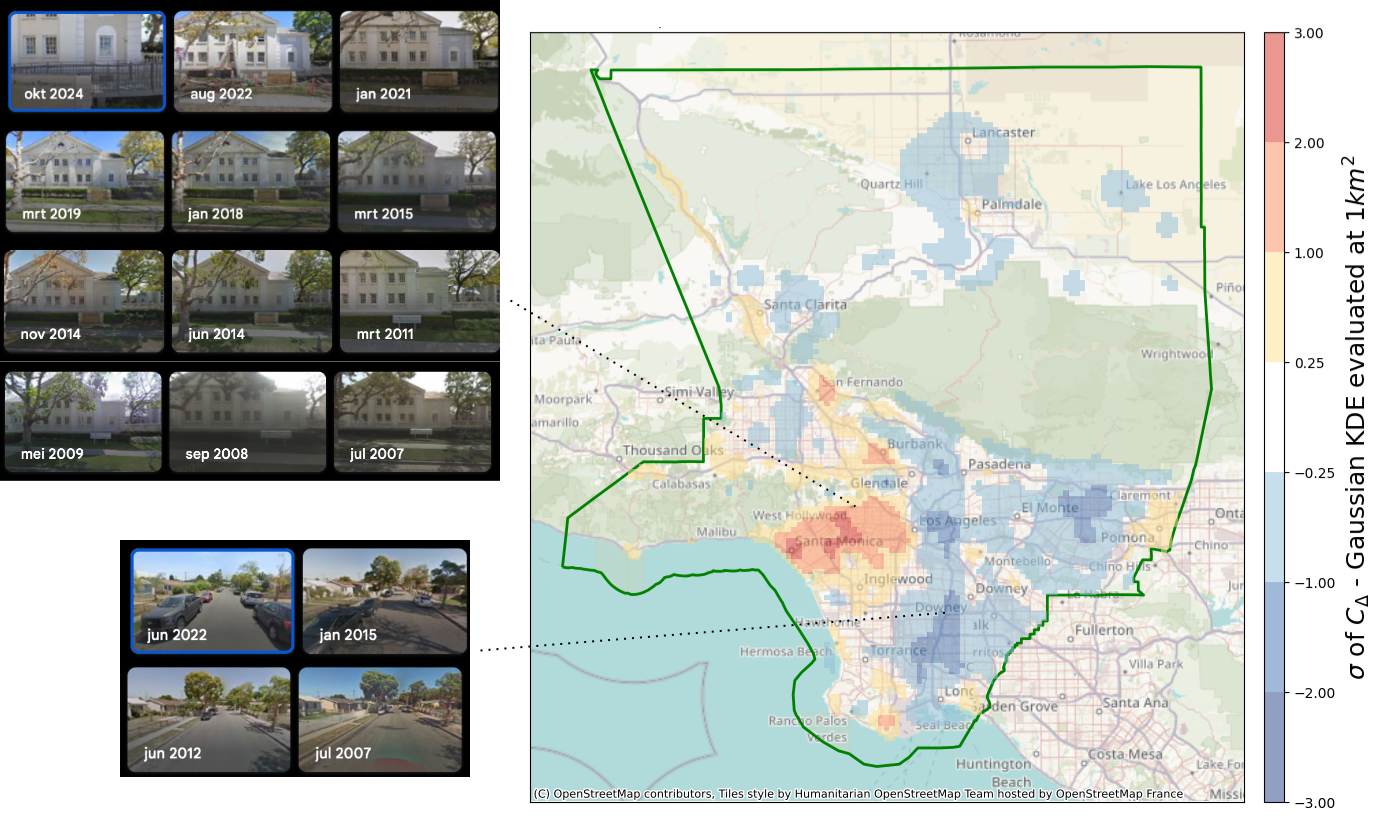}
    \caption{Density plot ($C_\Delta$) of Google Street View Coverage in Los Angeles. Oversampled neighbourhoods such as Beverly Hills can have 12-14 images in a suburban street whereas undersampled neighbourhoods such as Compton may only have 4-5 images on similar streets.}
    \label{fig:LA_Qualitative}
\end{figure*}
\subsection{Utility of Coverage Distribution}

\begin{figure*}
    \centering
    \begin{subfigure}[b]{0.45\textwidth}
        \centering
        \includegraphics[width=\textwidth]{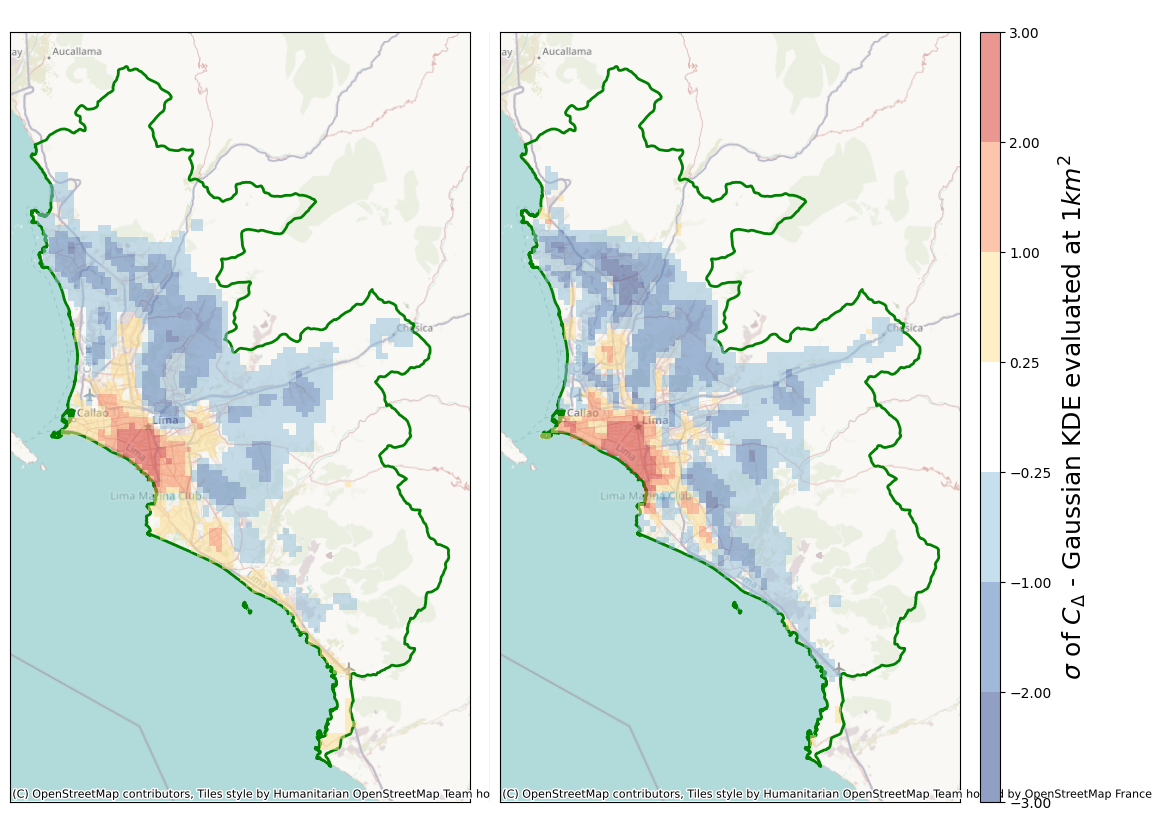}
        \caption{Density plots ($C_\Delta$) of Lima through Google Street View (left) and Mapillary (right).}
        \label{fig:cdelta_Lima}
    \end{subfigure}
    \hfill
    \begin{subfigure}[b]{0.5\textwidth}
        \centering
        \includegraphics[width=\textwidth]{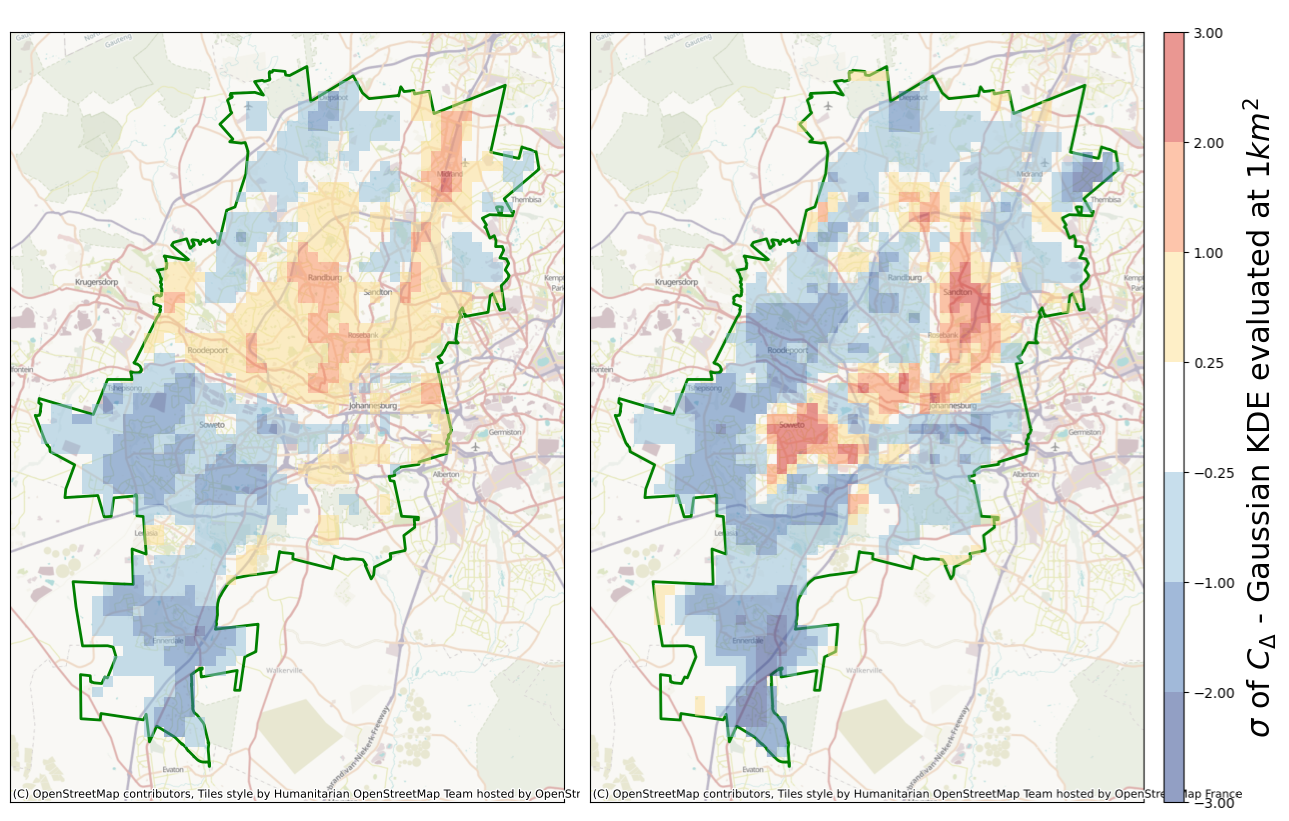}
        \caption{Density plots ($C_\Delta$) of Johannesburg through Google Street View (left) and Mapillary (right).}
        \label{fig:cdelta_johannesburg}
    \end{subfigure}
    \caption{Differences and similarities in coverage for Lima and Johannesburg. While Google Street View and Mapillary have similar coverage patterns in Lima, for Johannesburg it differs significantly.}
    \label{fig:Lima_and_Johannesburg}
\end{figure*}

To further evaluate the utility of the coverage distribution opposed to just using binary coverage, we plot the coverage percentage against the distance metrics for our chosen cities for both Google Street View and Mapillary. The results of this can be seen in Figure~\ref{fig:worldcities_covcorplot}. The plots of all other settings for KL, Mapillary and using all publicly accessible streets can be seen in the Appendix in Figure~\ref{fig:app_GSV_Drive_KL_firsttt}-\ref{fig:app_MLY_Pub_KL_lasttt}.

We observe that while there seems to be a slight correlation between the EMD and coverage percentage for Mapillary ($R^2 = 0.07)$, overall there is no correlation between coverage percentage and coverage distribution. As such we can conclude that measuring using coverage percentage alone is not sufficient to capture the way in which a city is covered.
Furthermore, we observe that coverage percentages are generally lower for cities on Mapillary than they are on GSV. For the coverage percentages across both providers we see that cities in the Global South tend to have less coverage than the Global North. However, this trend does not appear to repeat itself in terms of the coverage distribution. In Figure~\ref{fig:worldcities_covcorplot} we see how Tokyo, Auckland, and Los Angeles, large metropolitan areas in the Global North, have close to identical coverage percentage but vastly differing EMD scores, This indicates that while all their streets have been covered, in Los Angeles the coverage pattern is much more skewed to certain neighbourhoods than in Tokyo. In the Global South the same goes for the relation between coverage and EMD between Johannesburg and Dakar, or between Nairobi, Lima, and Lagos. 
We also observe that the EMD tends to be lower for GSV than for Mapillary, while the opposite holds for the KL. This is because the KL is more sensitive to disjoint distributions, which is often the case in Mapillary data as some roads are covered a lot, while other roads are not covered at all. The EMD is generally lower for Mapillary because their coverage is not skewed towards certain neighbourhoods: all neighbourhoods have equally poor coverage.

For the KL we see similar patterns: For GSV, cities such as Pittsburgh, Vancouver, and Paris have close to perfect coverage but vastly different KL scores with Paris having a KL close to that of Accra, a city that has only 40\% of its streets covered on GSV.

Lastly, we observe that, in general, smaller cities tend to have a better EMD. This is to be expected, as the EMD is influenced by the distance between over and undersampled areas; even if the coverage pattern is skewed to certain neighborhoods, these centers of mass are spatially closer to each other in smaller cities. However, we again see differences here between areas of similar sizes. Lagos (6734 $m^2$) and Lima (6615 $m^2$) are highly similar in area with varying EMD for GSV. We see that in larger metropolitan areas, a choice is made more often to drive much more often in certain neighbourhoods of the city, while other parts are neglected. 
\begin{figure*}
    \centering
    \includegraphics[width=.96\linewidth]{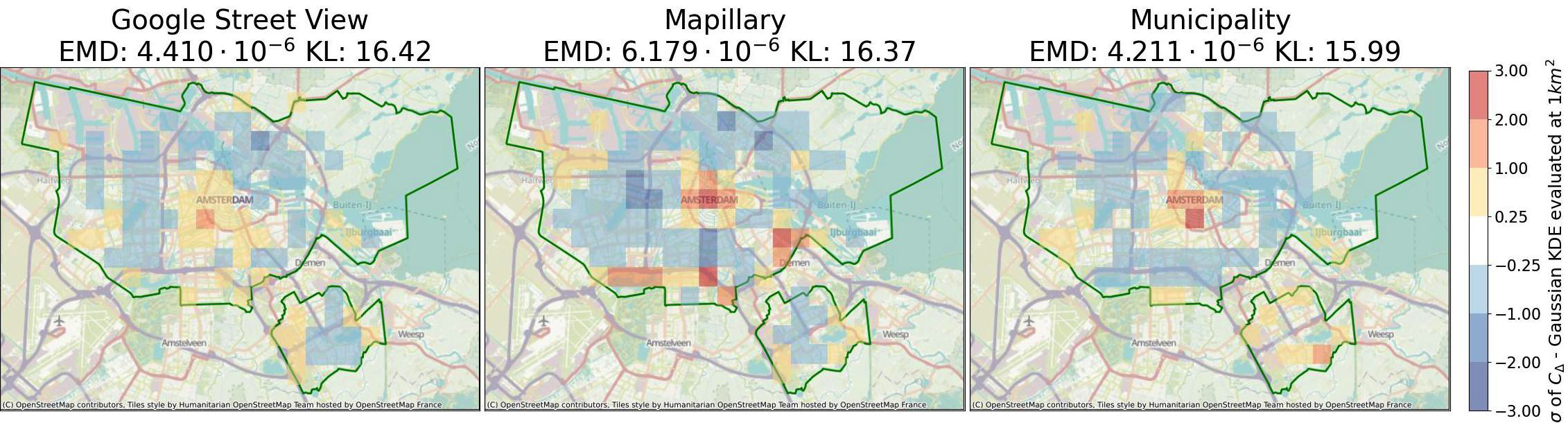}
    \caption{Density plots of $C_{\Delta}$ for coverage through Google Street View, Mapillary, and the Amsterdam Municipality. While the degree varies, similar patterns are observable for all three street view databases. The city center is oversampled, while the northern and eastern parts are undersampled.}
    \label{fig:Amsterdam_allproviders}
\end{figure*}
\subsection{Qualitative Analysis}
For further analysis we evaluate the difference maps $C_\Delta$ for all cities. We observe that this method of plotting the coverage distribution provides a comprehensive overview of coverage patterns. An example of the way coverage patterns can be observed is shown in Figure~\ref{fig:LA_Qualitative}. Here the $C_\Delta$ of Los Angeles for GSV is plotted. We can directly observe the coverage distribution where neighbourhoods such as Santa Monica, Beverly Hills, and West Hollywood are oversampled, while neighbourhoods such as Glendale, Compton, and West Covina are undersampled. Exploring through the GSV interfaces shows us the difference in images, with 12-14 images per suburban street in oversampled areas while a similar suburban street in an undersampled area might only have 4-5 images. These coverage patterns have previously been identified on a smaller scale, but had to be done so manually with a number of auditors. This approach allows us to identify patterns quickly at scale.

Another interesting observation is that the $C_\Delta$ for GSV and Mapillary can differ per city, indicating that coverage patterns might not always be a consequence of factors relating to the city, but due to the people driving around. In Figure~\ref{fig:cdelta_Lima} we see that the coverage patterns for Lima across both providers are strikingly similar. In Figure~\ref{fig:cdelta_johannesburg} we see two different patterns. While GSV oversamples in Midrand and Randburg, and undersamples in Soweto, this is swapped for Mapillary. The crowdsourced nature of Mapillary could potentially be a reason for this, as users can drive to collect data themselves they might be less inclined to follow predetermined patterns or be motivated to cover areas in the city that have not been adequately captured by existing street view providers. 

\begin{figure*}[]
    \centering
    \begin{minipage}{.56\textwidth}
    \centering
    \includegraphics[width=.9\linewidth]{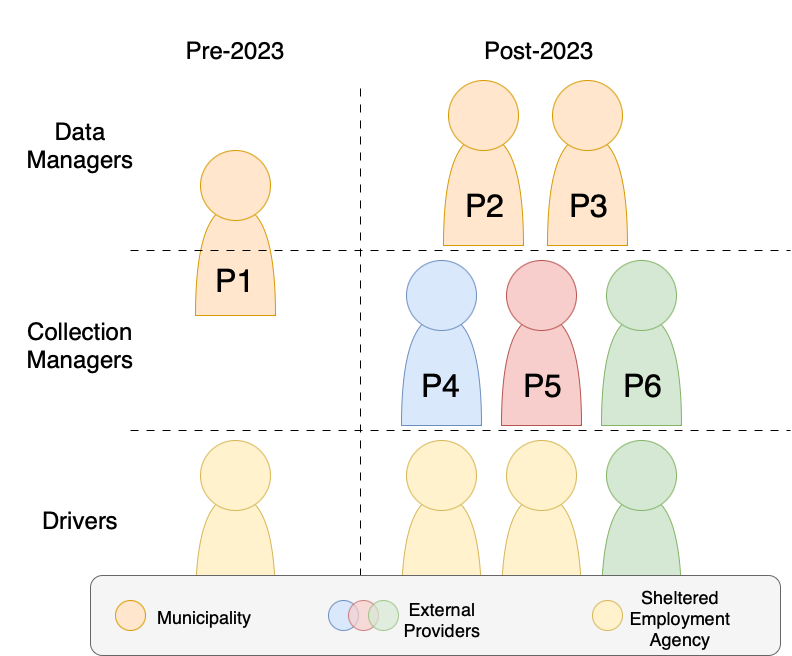}
    \caption{Diagram of workflow regarding the collection process of the Amsterdam Panorama Database. Pre-2023 the collection process was overseen directly by the municipality (Orange). In 2023 this process was outsourced to 3 different parties.}
    \label{fig:Intervieweelegend}
    \end{minipage}%
    \hfill
    \begin{minipage}{.43\textwidth}
    \centering
    \includegraphics[width=.9\linewidth]{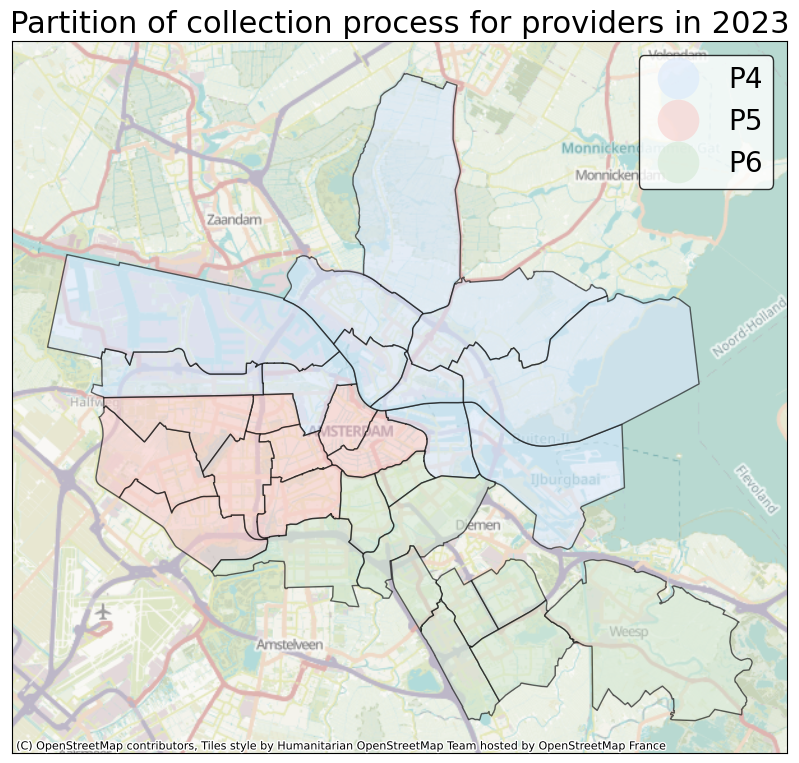}
    \caption{Disjointed areas for in which external providers collect street view images for the municipality of Amsterdam. }
    \label{fig:collection_areas_providers}
    \end{minipage}
\end{figure*}
\section{Case Study of Amsterdam}

To study the factors that influence the collection street view images we perform a case study of the panorama database of the municipality of Amsterdam.
This panorama database was first constructed in 2016 by the municipality as an effort to collect street view data throughout Amsterdam similarly to providers such as GSV and Mapillary. 
The Amsterdam municipality has the aim to cover the entire city in an equal manner, with no coverage differences between neighbourhoods, which is possible to evaluate using our method.

To evaluate the differences in coverage distributions for Amsterdam we perform an analysis with our proposed method. For this analysis we compare between the uniform and real coverage scenarios as well as to the coverage patterns of different providers. Amsterdam is highly suitable for a comparison between providers as it is the only city in our dataset that has more that 95\% of the streets covered for both GSV as well as Mapillary.
Moreover, we conduct semi-structured in-depth interviews with practitioners from both the municipality as well as external parties to gain a deeper understanding how the collection process itself introduces certain artifacts into the data. We started by contacting the data managers of the panoramas database, and we subsequently recruited further participants through \textit{snowball sampling}.

\subsection{Semi-structured interviews}
We conducted 6 semi-structured interviews each lasting for approximately 40 minutes. The interview questions were formulated in open-ended way to allow participants to share information in their own words while adhering to a general structure of topics \cite{fujii_interviewing_2018,Goede_secrecy_2019, Dankloff_2024}. The questions can be found in the Appendix in Table~\ref{interviewquestions}. The questions are designed to map the collection process and are divided into three sections: (1) Details regarding the driving patterns, (2) goals and requirements of the collection process, and (3) questions regarding the technical specifications of equipment and processing. 
Through snowball sampling we interviewed 6 people within the process chain. A diagram of the chain can be found in Figure~\ref{fig:Intervieweelegend}. We found that we could identify three clear roles within the process chain: Drivers that are responsible for the actual driving of the collection vehicle, Collection managers, that oversee the driving process and coordinate the collection process, and Datamanagers, that coordinate the collection process by managing the database. In our case the data managers are also referred to as \textit{owners}. 
The Amsterdam street view data from before 2023 was collected by the municipality itself, from 2023 onwards the collection process was outsourced to external parties. We will refer to these two phases as pre and post 2023. Pre-2023 the data manager communicated directly with the drivers while post-2023 this process has one extra step in it in the form of the collection managers of the external parties.

\subsection{Findings}
The interviews yielded multiple insights regarding the collection process. When asked how the decision is made where the street view car drives both P1 and P2 immediately put forward a principle of equality (\textit{``gelijkheidsbeginsel''}). The Amsterdam Panorama Database is constructed with this principle in mind. That every year, the entirety of Amsterdam needs to be covered. With no distinction made between neighbourhoods based on population, popularity, or any socio-economic factor. This principle has held both pre and post 2023, and is exercised in the form of a digital road log accessible by the drivers that registers what roads have been driven. 
Pre-2023, 3 external drivers from a sheltered employment agency drove the city using the driving route log. They were not micromanaged in what routes they drove, but the log was updated throughout the year so they could more systematically tackle different parts of the city in order. P1 mentioned that \textit{``obstructions can cause you to ride a certain road more often if you have to come back''}.

Post-2023 the collection process was outsourced to three providers: P4, P5, and P6. The collection process differs slightly for all three providers. All providers are responsible for a selected part of Amsterdam, this can be seen in Figure~\ref{fig:collection_areas_providers}. The disjointed areas allocated to each provider provide a potential gateway for bias if their approaches to collection differ.

All three providers (P4,P5,P6) mention that the city center is more "tricky" than other neighbourhoods due to obstructions or traffic. As such, P4 mentions they work with a set of drivers but always only one per car. They decide themselves based on their "emotional state", which refers to the mental fortitude for dealing with traffic to stay safe. If they can't keep a safe working environment they will continue collection outside the city center. This allows them to drive 90km per day. P5 explains they drive less per day (30-35km), and tries to work from outside in, to do the center in multiple passes. They work with 2 people per car, one to read the map and one to drive. P6 let's the driver decide, as it's \textit{``impossible to decide how drivers should drive top down''}. They work with 1 driver who is highly experienced.

In addition to the differences in how the collection is done, there also exist differences in the post-processing stage;
P4 performs the deduplication of spatially proximate images themselves, while P5 and P6 send all collected images to the municipality. There are also differences in the technical equipment used: P4, P5, and P6 use a 100MP, 48MP, and 72MP camera respectively. While this resolution difference may not be an issue for human observation, AI models could overfit on artifacts related to the resolution.

The time of year in which collection is done also varies per provider, with P4 collecting from August to October, P5 collecting from April to September, and P6 collecting from \textit{``spring until summer''}. The municipality requests that collection is done between March and October as collection depends on good weather, but does not specify this further. Moreover, collection has to be paused when it rains, and preferably no driving on clouded days. This is understandable, but as the city has been divided into disjoint sets this could result in seasonal/weather biases across areas. 

The main insights gained from these interviews are twofold: (1) The collection process is highly idiosyncratic, with individual collection car drivers being allowed to decide themselves how to drive. This differs from the perception that street view data is collected in a systematic way. Of all interviewees, P5 was the only person to mention that they would like to move to a more systematic way of collecting. (2) The interviews give insight into why we see similar patterns for each of the providers in Figure~\ref{fig:Amsterdam_allproviders}. As the nature of the city center forces drivers to return another day when confronted with obstructions or traffic this will result in oversampling in such areas if these images are not filtered out by spatial proximity. P1 explained that there is post-processing software to filter collected images before uploading to the database, but this only filters for spatial proximity when taken on the same day. If a driver encounters an obstruction and has to come back the next day images on the same location for both days will be uploaded. While density maps give the impression that the coverage distribution is skewed because of certain socio-economic factors we observe that in Amsterdam it are human idiosyncrasies that skew the data unknowingly.

\section{Limitations}
Our approach presents a novel way of evaluating the coverage of street view imagery, yet there are some factors which inhibit the comparison, which we will discuss in the following sections. 

\subsection{Dependency on Open Street Map}
The collection of street view data and calculation of metrics is reliant on data provided by OpenStreetMap. This is a useful resource as it allows us to collect data from a single place thus keeping a consistent experimental setup. However, there are a number of limitations that come with using OpenStreetMap.

Firstly, administrative city boundaries are relatively arbitrary concepts. When evaluating coverage this raises questions regarding where to cut the evaluation. For a clean experimental setup we choose to adhere to the administrative boundaries provided by OpenStreetMap, but they are not necessarily similar for all cities. Buenos Aires is cut off from the larger metropolitan area by its administrative boundary whereas the boundary of Bangkok includes a large part of the sea. Furthermore in OpenStreetMap, not all cities \textit{have} a polygon for the administrative boundaries. As such, performing similar evaluations for smaller cities globally would have to mix more local data sources. This adds to the argument that large generalized approaches for mining street view data potentially turn a blind eye to differences specific to some cities. 

Secondly, cities such as Nairobi or Singapore have a number of driveable roads in housing estates that are blocked off by gates . As such, roads in gated communities are rarely captured and form direct cause as to why the coverage distribution differs from street network returned by OpenStreetMap.
Furthermore, in cities such as Accra we observed that while larger access roads were included in the OSM network, the smaller roads they branched into weren't always present. Finally, OpenStreetMap in general uses crowdsourcing which could potentially make it a biased data source. 


\subsection{Dependency on coverage}

While our method uncovers coverage patterns for cities in which there is coverage, our method does not produce great results for cities in which there is little to no coverage. As such we recommend to evaluate coverage distribution only when there is a basis of initial coverage. As such, cities in North Africa were generally excluded from the research due to the lack of good coverage across both Google Street View as well as Mapillary. 

Secondly, across the cities we evaluated, the Mapillary coverage was significantly less than the Google Street View coverage. While this does not hold for all cities, the density distributions uncovered through Mapillary are more indicative of the spatial coverage itself than its distribution. 

Finally, while we uncover coverage patterns for some cities, we cannot say that certain coverage patterns hold for all cities. We also do not claim that cities in the Global North have ``better'' coverage than cities in the Global South, our findings show that coverage is dependent on idiosyncracies relating to the humans that collect the data as well as local infrastructure or city layout unique to the city. However, while these limitations are present in our research, we believe they do not invalidate our findings, and instead further support them. Because, due to the large differences between cities even within continents, a generalisable approach to learning patterns over cities is not favourable. We therefore believe that domain knowledge of the city and data sources are necessary to create datasets that are representative of the city. 

\subsection{Interviews}

A main limitation of our interviews is that we did interview any drivers directly. Due to the nature of the drivers' sheltered employment status both the municipality and the external providers requested we not interview them. We chose to not pursue this further as we are not trained to interview potentially vulnerable individuals. The one external provider employing a professional driver denied us access for personal reasons.

However, the differences in instructions provided by the managers may already explain some of the idiosyncrasies of the data. Their instructions vary in: distance covered per day, amount of drivers per car, systems for determining routes, in which months they collect, and their camera specifications. We do not intend to claim these findings generalize across cities, but we hope the insights from the interviews encourage other researchers to study driving patterns at a larger scale. Moreover, insight into the collection process in Amsterdam may aid in understanding Google Street View, as both use outside contractors and similar differences between contractors may occur.




\section{Conclusion}

We evaluated the coverage distribution of 28 cities globally. We determined the utility of distribution as a metric by comparing it to binary coverage and saw that only defining coverage in a binary way is insufficient to paint a picture of a well covered city. We further performed a case study of the Amsterdam panorama data through semi-structured interviews to better understand how the street view data collection is a human as opposed to mechanistic process. We found that street view data collection is influenced by idiosyncracies across top level policies, emotional state of drivers, and even city layout. As such we concluded that domain knowledge is necessary to account for the possible biases in the data when building AI datasets using imagery from street view databases. We hope these findings do not just allow for new sampling techniques to be developed to filter out these biases but also to address the root cause of biases in the collection process, and the human aspect that influences it.

\bibliographystyle{ACM-Reference-Format}
\bibliography{sample-base}


\begin{thebibliography}{58}


\ifx \showCODEN    \undefined \def \showCODEN     #1{\unskip}     \fi
\ifx \showISBNx    \undefined \def \showISBNx     #1{\unskip}     \fi
\ifx \showISBNxiii \undefined \def \showISBNxiii  #1{\unskip}     \fi
\ifx \showISSN     \undefined \def \showISSN      #1{\unskip}     \fi
\ifx \showLCCN     \undefined \def \showLCCN      #1{\unskip}     \fi
\ifx \shownote     \undefined \def \shownote      #1{#1}          \fi
\ifx \showarticletitle \undefined \def \showarticletitle #1{#1}   \fi
\ifx \showURL      \undefined \def \showURL       {\relax}        \fi
\providecommand\bibfield[2]{#2}
\providecommand\bibinfo[2]{#2}
\providecommand\natexlab[1]{#1}
\providecommand\showeprint[2][]{arXiv:#2}

\bibitem[Alcantarilla et~al\mbox{.}(2016)]%
        {Alcantarilla2016}
\bibfield{author}{\bibinfo{person}{Pablo~F. Alcantarilla}, \bibinfo{person}{Simon Stent}, \bibinfo{person}{German Ros}, \bibinfo{person}{Roberto Arroyo}, {and} \bibinfo{person}{Riccardo Gherardi}.} \bibinfo{year}{2016}\natexlab{}.
\newblock \showarticletitle{Street-View Change Detection with Deconvolutional Networks}.
\newblock \bibinfo{journal}{\emph{Robotics: Science and Systems}}  \bibinfo{volume}{12} (\bibinfo{year}{2016}).
\newblock
\showISBNx{9780992374723}
\showISSN{2330765X}
\href{https://doi.org/10.15607/rss.2016.xii.044}{doi:\nolinkurl{10.15607/rss.2016.xii.044}}


\bibitem[Ali-bey et~al\mbox{.}(2022)]%
        {Ali_bey_2022}
\bibfield{author}{\bibinfo{person}{Amar Ali-bey}, \bibinfo{person}{Brahim Chaib-draa}, {and} \bibinfo{person}{Philippe Giguère}.} \bibinfo{year}{2022}\natexlab{}.
\newblock \showarticletitle{GSV-Cities: Toward appropriate supervised visual place recognition}.
\newblock \bibinfo{journal}{\emph{Neurocomputing}}  \bibinfo{volume}{513} (\bibinfo{date}{Nov.} \bibinfo{year}{2022}), \bibinfo{pages}{194–203}.
\newblock
\showISSN{0925-2312}
\href{https://doi.org/10.1016/j.neucom.2022.09.127}{doi:\nolinkurl{10.1016/j.neucom.2022.09.127}}


\bibitem[Alpherts et~al\mbox{.}(2024)]%
        {alphertsPerceptiveVisualUrban2024}
\bibfield{author}{\bibinfo{person}{Tim Alpherts}, \bibinfo{person}{Sennay Ghebreab}, \bibinfo{person}{Yen-Chia Hsu}, {and} \bibinfo{person}{Nanne Van~Noord}.} \bibinfo{year}{2024}\natexlab{}.
\newblock \showarticletitle{Perceptive {{Visual Urban Analytics}} Is {{Not}} ({{Yet}}) {{Suitable}} for {{Municipalities}}}. In \bibinfo{booktitle}{\emph{The 2024 {{ACM Conference}} on {{Fairness}}, {{Accountability}}, and {{Transparency}}}}. \bibinfo{publisher}{ACM}, \bibinfo{address}{Rio de Janeiro Brazil}, \bibinfo{pages}{1341--1354}.
\newblock
\showISBNx{9798400704505}
\href{https://doi.org/10.1145/3630106.3658976}{doi:\nolinkurl{10.1145/3630106.3658976}}


\bibitem[Arandjelovic et~al\mbox{.}(2018)]%
        {Arandjelovi}
\bibfield{author}{\bibinfo{person}{Relja Arandjelovic}, \bibinfo{person}{Petr Gronat}, \bibinfo{person}{Akihiko Torii}, \bibinfo{person}{Tomas Pajdla}, {and} \bibinfo{person}{Josef Sivic}.} \bibinfo{year}{2018}\natexlab{}.
\newblock \showarticletitle{{{NetVLAD}}: {{CNN Architecture}} for {{Weakly Supervised Place Recognition}}}.
\newblock \bibinfo{journal}{\emph{IEEE Transactions on Pattern Analysis and Machine Intelligence}} \bibinfo{volume}{40}, \bibinfo{number}{6} (\bibinfo{year}{2018}), \bibinfo{pages}{1437--1451}.
\newblock
\showISSN{01628828}
\href{https://doi.org/10.1109/TPAMI.2017.2711011}{doi:\nolinkurl{10.1109/TPAMI.2017.2711011}}
\showeprint[arxiv]{1511.07247}
\newblock
\shownote{NetVLAD, what else?}.


\bibitem[Arietta et~al\mbox{.}(2014)]%
        {Arietta2014}
\bibfield{author}{\bibinfo{person}{Sean~M. Arietta}, \bibinfo{person}{Alexei~A. Efros}, \bibinfo{person}{Ravi Ramamoorthi}, {and} \bibinfo{person}{Maneesh Agrawala}.} \bibinfo{year}{2014}\natexlab{}.
\newblock \showarticletitle{City Forensics: {{Using}} Visual Elements to Predict Non-Visual City Attributes}.
\newblock \bibinfo{journal}{\emph{IEEE Transactions on Visualization and Computer Graphics}} \bibinfo{volume}{20}, \bibinfo{number}{12} (\bibinfo{year}{2014}), \bibinfo{pages}{2624--2633}.
\newblock
\showISSN{10772626}
\href{https://doi.org/10.1109/TVCG.2014.2346446}{doi:\nolinkurl{10.1109/TVCG.2014.2346446}}


\bibitem[Berton et~al\mbox{.}(2022)]%
        {berton2022rethinkingvisualgeolocalizationlargescale}
\bibfield{author}{\bibinfo{person}{Gabriele Berton}, \bibinfo{person}{Carlo Masone}, {and} \bibinfo{person}{Barbara Caputo}.} \bibinfo{year}{2022}\natexlab{}.
\newblock \bibinfo{title}{Rethinking Visual Geo-localization for Large-Scale Applications}.
\newblock
\showeprint[arxiv]{2204.02287}~[cs.CV]
\urldef\tempurl%
\url{https://arxiv.org/abs/2204.02287}
\showURL{%
\tempurl}


\bibitem[Camara et~al\mbox{.}(2019)]%
        {Camara2019}
\bibfield{author}{\bibinfo{person}{Luis~G Camara}, \bibinfo{person}{Carl G{\"a}bert}, {and} \bibinfo{person}{Libor P{\textasciicaron}}.} \bibinfo{year}{2019}\natexlab{}.
\newblock \showarticletitle{Highly {{Robust Visual Place Recognition Through Spatial Matching}} of {{CNN Highly Robust Visual Place Recognition Through Spatial Matching}} of {{CNN Features}}}.
\newblock  \bibinfo{number}{September} (\bibinfo{year}{2019}).
\newblock
\newblock
\shownote{Refers to Pittsburgh 30k}.


\bibitem[Cao and Snavely(2015)]%
        {Cao2015}
\bibfield{author}{\bibinfo{person}{Song Cao} {and} \bibinfo{person}{Noah Snavely}.} \bibinfo{year}{2015}\natexlab{}.
\newblock \showarticletitle{Graph-{{Based Discriminative Learning}} for {{Location Recognition}}}.
\newblock \bibinfo{journal}{\emph{International Journal of Computer Vision}} \bibinfo{volume}{112}, \bibinfo{number}{2} (\bibinfo{year}{2015}), \bibinfo{pages}{239--254}.
\newblock
\showISSN{15731405}
\href{https://doi.org/10.1007/s11263-014-0774-9}{doi:\nolinkurl{10.1007/s11263-014-0774-9}}
\newblock
\shownote{Represented visual place recognition database structure in a graph. Still only using bow-based location recognition.}.


\bibitem[Che et~al\mbox{.}(2019)]%
        {che2019d2citylargescaledashcamvideo}
\bibfield{author}{\bibinfo{person}{Zhengping Che}, \bibinfo{person}{Guangyu Li}, \bibinfo{person}{Tracy Li}, \bibinfo{person}{Bo Jiang}, \bibinfo{person}{Xuefeng Shi}, \bibinfo{person}{Xinsheng Zhang}, \bibinfo{person}{Ying Lu}, \bibinfo{person}{Guobin Wu}, \bibinfo{person}{Yan Liu}, {and} \bibinfo{person}{Jieping Ye}.} \bibinfo{year}{2019}\natexlab{}.
\newblock \bibinfo{title}{D$^2$-City: A Large-Scale Dashcam Video Dataset of Diverse Traffic Scenarios}.
\newblock
\showeprint[arxiv]{1904.01975}~[cs.LG]
\urldef\tempurl%
\url{https://arxiv.org/abs/1904.01975}
\showURL{%
\tempurl}


\bibitem[Curtis et~al\mbox{.}(2013)]%
        {curtisUsingGoogleStreet2013}
\bibfield{author}{\bibinfo{person}{Jacqueline~W. Curtis}, \bibinfo{person}{Andrew Curtis}, \bibinfo{person}{Jennifer Mapes}, \bibinfo{person}{Andrea~B. Szell}, {and} \bibinfo{person}{Adam Cinderich}.} \bibinfo{year}{2013}\natexlab{}.
\newblock \showarticletitle{Using Google Street View for Systematic Observation of the Built Environment: Analysis of Spatio-Temporal Instability of Imagery Dates}.
\newblock  \bibinfo{volume}{12}, \bibinfo{number}{1} (\bibinfo{year}{2013}), \bibinfo{pages}{53}.
\newblock
\showISSN{1476-072X}
\href{https://doi.org/10.1186/1476-072X-12-53}{doi:\nolinkurl{10.1186/1476-072X-12-53}}


\bibitem[Dankloff et~al\mbox{.}(2024)]%
        {Dankloff_2024}
\bibfield{author}{\bibinfo{person}{Mirthe Dankloff}, \bibinfo{person}{Vanja Skoric}, \bibinfo{person}{Giovanni Sileno}, \bibinfo{person}{Sennay Ghebreab}, \bibinfo{person}{Jacco~van Ossenbruggen}, {and} \bibinfo{person}{Emma Beauxis-Aussalet}.} \bibinfo{year}{2024}\natexlab{}.
\newblock \showarticletitle{Analysing and organising human communications for AI fairness assessment: Use cases from the Dutch Public Sector}.
\newblock \bibinfo{journal}{\emph{AI \& SOCIETY}} (\bibinfo{date}{June} \bibinfo{year}{2024}).
\newblock
\showISSN{1435-5655}
\href{https://doi.org/10.1007/s00146-024-01974-4}{doi:\nolinkurl{10.1007/s00146-024-01974-4}}


\bibitem[De~Goede et~al\mbox{.}(2019)]%
        {Goede_secrecy_2019}
\bibfield{author}{\bibinfo{person}{Marieke De~Goede}, \bibinfo{person}{Bosma Esmé}, {and} \bibinfo{person}{Pallister-Wilkins Polly}.} \bibinfo{year}{2019}\natexlab{}.
\newblock \bibinfo{booktitle}{\emph{Secrecy and Methods in Security Research A Guide to Qualitative Fieldwork.}}
\newblock \bibinfo{publisher}{Routledge}, \bibinfo{address}{New York, NY ; Abingdon, Oxon}.
\newblock
\showISBNx{9780367027247}
\urldef\tempurl%
\url{https://www.routledge.com/Secrecy-and-Methods-in-Security-Research-A-Guide-to-Qualitative-Fieldwork/DeGoede-Bosma-Pallister-Wilkins/p/book/9780367027247?srsltid=AfmBOop46NsJPo73q8w8ME_QDW1xCbTvvl4zNEeE0t7wVsIpho7mIY-I}
\showURL{%
\tempurl}


\bibitem[Dubey et~al\mbox{.}(2016)]%
        {Dubey2016}
\bibfield{author}{\bibinfo{person}{A Dubey}, \bibinfo{person}{N Nikhil}, \bibinfo{person}{D Parikh}, \bibinfo{person}{R Raskar}, {and} \bibinfo{person}{C{\'e}sar~A. Hidalgo}.} \bibinfo{year}{2016}\natexlab{}.
\newblock \showarticletitle{Deep {{Learning}} the {{City}}: {{Quantifying Urban Perception}} at a {{Global Scale}}}.
\newblock \bibinfo{journal}{\emph{Eccv}}  \bibinfo{volume}{3} (\bibinfo{year}{2016}), \bibinfo{pages}{398--413}.
\newblock
\showISBNx{9783319464480}
\href{https://doi.org/10.1007/978-3-319-46448-0}{doi:\nolinkurl{10.1007/978-3-319-46448-0}}


\bibitem[Ertler et~al\mbox{.}(2020)]%
        {ertler2020mapillarytrafficsigndataset}
\bibfield{author}{\bibinfo{person}{Christian Ertler}, \bibinfo{person}{Jerneja Mislej}, \bibinfo{person}{Tobias Ollmann}, \bibinfo{person}{Lorenzo Porzi}, \bibinfo{person}{Gerhard Neuhold}, {and} \bibinfo{person}{Yubin Kuang}.} \bibinfo{year}{2020}\natexlab{}.
\newblock \bibinfo{title}{The Mapillary Traffic Sign Dataset for Detection and Classification on a Global Scale}.
\newblock
\showeprint[arxiv]{1909.04422}~[cs.CV]
\urldef\tempurl%
\url{https://arxiv.org/abs/1909.04422}
\showURL{%
\tempurl}


\bibitem[Feydy et~al\mbox{.}(2018)]%
        {feydy2018interpolatingoptimaltransportmmd}
\bibfield{author}{\bibinfo{person}{Jean Feydy}, \bibinfo{person}{Thibault Séjourné}, \bibinfo{person}{François-Xavier Vialard}, \bibinfo{person}{Shun ichi Amari}, \bibinfo{person}{Alain Trouvé}, {and} \bibinfo{person}{Gabriel Peyré}.} \bibinfo{year}{2018}\natexlab{}.
\newblock \bibinfo{title}{Interpolating between Optimal Transport and MMD using Sinkhorn Divergences}.
\newblock
\showeprint[arxiv]{1810.08278}~[math.ST]
\urldef\tempurl%
\url{https://arxiv.org/abs/1810.08278}
\showURL{%
\tempurl}


\bibitem[Franchi et~al\mbox{.}(2023)]%
        {Franchi_2023}
\bibfield{author}{\bibinfo{person}{Matt Franchi}, \bibinfo{person}{J.D. Zamfirescu-Pereira}, \bibinfo{person}{Wendy Ju}, {and} \bibinfo{person}{Emma Pierson}.} \bibinfo{year}{2023}\natexlab{}.
\newblock \showarticletitle{Detecting disparities in police deployments using dashcam data}. In \bibinfo{booktitle}{\emph{2023 ACM Conference on Fairness, Accountability, and Transparency}} \emph{(\bibinfo{series}{FAccT ’23})}. \bibinfo{publisher}{ACM}, \bibinfo{pages}{534–544}.
\newblock
\href{https://doi.org/10.1145/3593013.3594020}{doi:\nolinkurl{10.1145/3593013.3594020}}


\bibitem[Fry et~al\mbox{.}(2020)]%
        {fryAssessingGoogleStreet2020}
\bibfield{author}{\bibinfo{person}{Dustin Fry}, \bibinfo{person}{Stephen~J. Mooney}, \bibinfo{person}{Daniel~A. Rodr{\'i}guez}, \bibinfo{person}{Waleska~T. Caiaffa}, {and} \bibinfo{person}{Gina~S. Lovasi}.} \bibinfo{year}{2020}\natexlab{}.
\newblock \showarticletitle{Assessing {{Google Street View Image Availability}} in {{Latin American Cities}}}.
\newblock \bibinfo{journal}{\emph{Journal of Urban Health}} \bibinfo{volume}{97}, \bibinfo{number}{4} (\bibinfo{date}{Aug.} \bibinfo{year}{2020}), \bibinfo{pages}{552--560}.
\newblock
\showISSN{1468-2869}
\href{https://doi.org/10.1007/s11524-019-00408-7}{doi:\nolinkurl{10.1007/s11524-019-00408-7}}


\bibitem[Fu et~al\mbox{.}(2018)]%
        {Fu2018}
\bibfield{author}{\bibinfo{person}{Kaiqun Fu}, \bibinfo{person}{Zhiqian Chen}, {and} \bibinfo{person}{Chang~Tien Lu}.} \bibinfo{year}{2018}\natexlab{}.
\newblock \showarticletitle{{{StreetNet}}: {{Preference}} Learning with Convolutional Neural Network on Urban Crime Perception}.
\newblock \bibinfo{journal}{\emph{GIS: Proceedings of the ACM International Symposium on Advances in Geographic Information Systems}} \bibinfo{number}{August 2019} (\bibinfo{year}{2018}), \bibinfo{pages}{269--278}.
\newblock
\showISBNx{9781450358897}
\href{https://doi.org/10.1145/3274895.3274975}{doi:\nolinkurl{10.1145/3274895.3274975}}


\bibitem[Fujii(2018)]%
        {fujii_interviewing_2018}
\bibfield{author}{\bibinfo{person}{Lee~Ann Fujii}.} \bibinfo{year}{2018}\natexlab{}.
\newblock \bibinfo{booktitle}{\emph{Interviewing in {Social} {Science} {Research}: a {Relational} {Approach}}}.
\newblock \bibinfo{publisher}{Routledge}, \bibinfo{address}{New York, NY ; Abingdon, Oxon}.
\newblock
\showISBNx{978-0-415-84372-0 978-0-415-84374-4}
\urldef\tempurl%
\url{https://www.routledge.com/Interviewing-in-Social-Science-Research-A-Relational-Approach/Fujii/p/book/9780415843744}
\showURL{%
\tempurl}


\bibitem[{Gauteng City-Region Observatory (GCRO)}(2016)]%
        {gcro2016qol4}
\bibfield{author}{\bibinfo{person}{{Gauteng City-Region Observatory (GCRO)}}.} \bibinfo{year}{2016}\natexlab{}.
\newblock \bibinfo{booktitle}{\emph{Quality of Life Survey 2015-2016 [dataset]. Round 4. Version 1}}.
\newblock Johannesburg; Cape Town.
\newblock
\href{https://doi.org/10.25828/w490-a496}{doi:\nolinkurl{10.25828/w490-a496}}


\bibitem[{Gauteng City-Region Observatory (GCRO)}(2021)]%
        {gcro2021qol6}
\bibfield{author}{\bibinfo{person}{{Gauteng City-Region Observatory (GCRO)}}.} \bibinfo{year}{2021}\natexlab{}.
\newblock \bibinfo{booktitle}{\emph{Quality of Life Survey 2020-2021 [dataset]. Round 6. Version 1}}.
\newblock Johannesburg; Cape Town.
\newblock
\href{https://doi.org/10.25828/wemz-vf31}{doi:\nolinkurl{10.25828/wemz-vf31}}


\bibitem[{Gauteng City-Region Observatory (GCRO)}(2022)]%
        {gcro2022qol5}
\bibfield{author}{\bibinfo{person}{{Gauteng City-Region Observatory (GCRO)}}.} \bibinfo{year}{2022}\natexlab{}.
\newblock \bibinfo{booktitle}{\emph{Quality of Life Survey V 2017-2018 [dataset]. Round 5. Version 2}}.
\newblock Johannesburg; Cape Town.
\newblock
\href{https://doi.org/10.25828/8yf7-9261}{doi:\nolinkurl{10.25828/8yf7-9261}}


\bibitem[Gebru et~al\mbox{.}(2017)]%
        {Gebru2017}
\bibfield{author}{\bibinfo{person}{Timnit Gebru}, \bibinfo{person}{Jonathan Krause}, \bibinfo{person}{Yilun Wang}, \bibinfo{person}{Duyun Chen}, \bibinfo{person}{Jia Deng}, \bibinfo{person}{Erez~Lieberman Aiden}, {and} \bibinfo{person}{Li {Fei-Fei}}.} \bibinfo{year}{2017}\natexlab{}.
\newblock \showarticletitle{Using Deep Learning and Google Street View to Estimate the Demographic Makeup of Neighborhoods across the {{United States}}}.
\newblock \bibinfo{journal}{\emph{Proceedings of the National Academy of Sciences of the United States of America}} \bibinfo{volume}{114}, \bibinfo{number}{50} (\bibinfo{year}{2017}), \bibinfo{pages}{13108--13113}.
\newblock
\showISSN{10916490}
\href{https://doi.org/10.1073/pnas.1700035114}{doi:\nolinkurl{10.1073/pnas.1700035114}}


\bibitem[{Gomez-Ojeda} et~al\mbox{.}(2015)]%
        {Gomez-Ojeda2015}
\bibfield{author}{\bibinfo{person}{Ruben {Gomez-Ojeda}}, \bibinfo{person}{Manuel {Lopez-Antequera}}, \bibinfo{person}{Nicolai Petkov}, {and} \bibinfo{person}{Javier {Gonzalez-Jimenez}}.} \bibinfo{year}{2015}\natexlab{}.
\newblock \showarticletitle{Training a {{Convolutional Neural Network}} for {{Appearance-Invariant Place Recognition}}}.
\newblock  (\bibinfo{year}{2015}), \bibinfo{pages}{1--9}.
\newblock
\showeprint[arxiv]{1505.07428}


\bibitem[Groenen et~al\mbox{.}(2022)]%
        {Groenen2022}
\bibfield{author}{\bibinfo{person}{Inske Groenen}, \bibinfo{person}{Stevan Rudinac}, {and} \bibinfo{person}{Marcel Worring}.} \bibinfo{year}{2022}\natexlab{}.
\newblock \showarticletitle{{{PanorAMS}}: {{Automatic Annotation}} for {{Detecting Objects}} in {{Urban Context}}}.
\newblock  (\bibinfo{year}{2022}).
\newblock
\showeprint[arxiv]{2208.14295}


\bibitem[Hamann(2024)]%
        {hamann2024segregation}
\bibfield{author}{\bibinfo{person}{Christiaan Hamann}.} \bibinfo{year}{2024}\natexlab{}.
\newblock \showarticletitle{Segregation and socio-economic sorting in Gauteng}.
\newblock \bibinfo{journal}{\emph{Map of the Month. Gauteng City-Region Observatory}} (\bibinfo{date}{August} \bibinfo{year}{2024}).
\newblock
\href{https://doi.org/10.36634/MIJV2656}{doi:\nolinkurl{10.36634/MIJV2656}}


\bibitem[Huang et~al\mbox{.}(2024)]%
        {huangCityPulseFineGrainedAssessment2024}
\bibfield{author}{\bibinfo{person}{Tianyuan Huang}, \bibinfo{person}{Zejia Wu}, \bibinfo{person}{Jiajun Wu}, \bibinfo{person}{Jackelyn Hwang}, {and} \bibinfo{person}{Ram Rajagopal}.} \bibinfo{year}{2024}\natexlab{}.
\newblock \bibinfo{title}{{{CityPulse}}: {{Fine-Grained Assessment}} of {{Urban Change}} with {{Street View Time Series}}}.
\newblock
\showeprint[arxiv]{2401.01107}~[cs]
\newblock
\shownote{Comment: Accepted by AAAI 2024}.


\bibitem[Ibrahimi et~al\mbox{.}(2021)]%
        {iovpr}
\bibfield{author}{\bibinfo{person}{Sarah Ibrahimi}, \bibinfo{person}{Nanne van Noord}, \bibinfo{person}{Tim Alpherts}, {and} \bibinfo{person}{Marcel Worring}.} \bibinfo{year}{2021}\natexlab{}.
\newblock \showarticletitle{Inside Out Visual Place Recognition}.
\newblock \bibinfo{journal}{\emph{CoRR}}  \bibinfo{volume}{abs/2111.13546} (\bibinfo{year}{2021}).
\newblock
\showeprint[arXiv]{2111.13546}
\urldef\tempurl%
\url{https://arxiv.org/abs/2111.13546}
\showURL{%
\tempurl}


\bibitem[Jacobs et~al\mbox{.}(2009)]%
        {jacobs09webcamgis}
\bibfield{author}{\bibinfo{person}{Nathan Jacobs}, \bibinfo{person}{Walker Burgin}, \bibinfo{person}{Nick Fridrich}, \bibinfo{person}{Austin Abrams}, \bibinfo{person}{Kylia Miskell}, \bibinfo{person}{Bobby~H. Braswell}, \bibinfo{person}{Andrew~D. Richardson}, {and} \bibinfo{person}{Robert Pless}.} \bibinfo{year}{2009}\natexlab{}.
\newblock \showarticletitle{The Global Network of Outdoor Webcams: Properties and Applications}. In \bibinfo{booktitle}{\emph{ACM SIGSPATIAL International Conference on Advances in Geographic Information Systems (ACM SIGSPATIAL)}}. \bibinfo{pages}{111--120}.
\newblock
\href{https://doi.org/10.1145/1653771.1653789}{doi:\nolinkurl{10.1145/1653771.1653789}}
\newblock
\shownote{Acceptance rate: 20.9\%}.


\bibitem[Kim and Jang(2023)]%
        {kimExaminationSpatialCoverage2023}
\bibfield{author}{\bibinfo{person}{Junghwan Kim} {and} \bibinfo{person}{Kee~Moon Jang}.} \bibinfo{year}{2023}\natexlab{}.
\newblock \showarticletitle{An Examination of the Spatial Coverage and Temporal Variability of {{Google Street View}} ({{GSV}}) Images in Small- and Medium-Sized Cities: {{A}} People-Based Approach}.
\newblock \bibinfo{journal}{\emph{Computers, Environment and Urban Systems}}  \bibinfo{volume}{102} (\bibinfo{date}{June} \bibinfo{year}{2023}), \bibinfo{pages}{101956}.
\newblock
\showISSN{0198-9715}
\href{https://doi.org/10.1016/j.compenvurbsys.2023.101956}{doi:\nolinkurl{10.1016/j.compenvurbsys.2023.101956}}


\bibitem[Law et~al\mbox{.}(2019)]%
        {Law2019}
\bibfield{author}{\bibinfo{person}{Stephen Law}, \bibinfo{person}{Brooks Paige}, {and} \bibinfo{person}{Chris Russell}.} \bibinfo{year}{2019}\natexlab{}.
\newblock \showarticletitle{Take a Look around: {{Using}} Street View and Satellite Images to Estimate House Prices}.
\newblock \bibinfo{journal}{\emph{ACM Transactions on Intelligent Systems and Technology}} \bibinfo{volume}{10}, \bibinfo{number}{5} (\bibinfo{year}{2019}), \bibinfo{pages}{1--19}.
\newblock
\showISSN{21576912}
\href{https://doi.org/10.1145/3342240}{doi:\nolinkurl{10.1145/3342240}}
\showeprint[arxiv]{1807.07155}


\bibitem[Ma et~al\mbox{.}(2022)]%
        {Ma2022}
\bibfield{author}{\bibinfo{person}{Nachuan Ma}, \bibinfo{person}{Jiahe Fan}, \bibinfo{person}{Wenshuo Wang}, \bibinfo{person}{Jin Wu}, \bibinfo{person}{Yu Jiang}, \bibinfo{person}{Lihua Xie}, {and} \bibinfo{person}{Rui Fan}.} \bibinfo{year}{2022}\natexlab{}.
\newblock \showarticletitle{Computer {{Vision}} for {{Road Imaging}} and {{Pothole Detection}}: {{A State-of-the-Art Review}} of {{Systems}} and {{Algorithms}}}.
\newblock  (\bibinfo{year}{2022}), \bibinfo{pages}{1--16}.
\newblock
\href{https://doi.org/10.1093/tse/tdac026}{doi:\nolinkurl{10.1093/tse/tdac026}}
\showeprint[arxiv]{2204.13590}


\bibitem[Machicao et~al\mbox{.}(2022)]%
        {machicao2022}
\bibfield{author}{\bibinfo{person}{Jeaneth Machicao}, \bibinfo{person}{Alison Specht}, \bibinfo{person}{Danton Vellenich}, \bibinfo{person}{Leandro Meneguzzi}, \bibinfo{person}{Romain David}, \bibinfo{person}{Shelley Stall}, \bibinfo{person}{Katia Ferraz}, \bibinfo{person}{Laurence Mabile}, \bibinfo{person}{Margaret O'brien}, {and} \bibinfo{person}{Pedro Corr{\^e}a}.} \bibinfo{year}{2022}\natexlab{}.
\newblock \showarticletitle{{A Deep-Learning Method for the Prediction of Socio-Economic Indicators from Street-View Imagery Using a Case Study from Brazil}}.
\newblock \bibinfo{journal}{\emph{{CODATA Data Science Journal}}}  \bibinfo{volume}{21} (\bibinfo{date}{Feb.} \bibinfo{year}{2022}).
\newblock
\href{https://doi.org/10.5334/dsj-2022-006}{doi:\nolinkurl{10.5334/dsj-2022-006}}


\bibitem[Miranda et~al\mbox{.}(2020)]%
        {Miranda2020}
\bibfield{author}{\bibinfo{person}{Fabio Miranda}, \bibinfo{person}{Maryam Hosseini}, \bibinfo{person}{Marcos Lage}, \bibinfo{person}{Harish Doraiswamy}, \bibinfo{person}{Graham Dove}, {and} \bibinfo{person}{Cl{\'a}udio~T. Silva}.} \bibinfo{year}{2020}\natexlab{}.
\newblock \showarticletitle{Urban {{Mosaic}}: {{Visual Exploration}} of {{Streetscapes Using Large-Scale Image Data}}}.
\newblock \bibinfo{journal}{\emph{Conference on Human Factors in Computing Systems - Proceedings}} (\bibinfo{year}{2020}).
\newblock
\showISBNx{9781450367080}
\href{https://doi.org/10.1145/3313831.3376399}{doi:\nolinkurl{10.1145/3313831.3376399}}
\showeprint[arxiv]{2008.13321}


\bibitem[Moreno~Berton et~al\mbox{.}(2021)]%
        {Moreno_Berton_2021}
\bibfield{author}{\bibinfo{person}{Gabriele Moreno~Berton}, \bibinfo{person}{Valerio Paolicelli}, \bibinfo{person}{Carlo Masone}, {and} \bibinfo{person}{Barbara Caputo}.} \bibinfo{year}{2021}\natexlab{}.
\newblock \showarticletitle{Adaptive-Attentive Geolocalization from few queries: a hybrid approach}. In \bibinfo{booktitle}{\emph{2021 IEEE Winter Conference on Applications of Computer Vision (WACV)}}. \bibinfo{publisher}{IEEE}, \bibinfo{pages}{2917–2926}.
\newblock
\href{https://doi.org/10.1109/wacv48630.2021.00296}{doi:\nolinkurl{10.1109/wacv48630.2021.00296}}


\bibitem[Moura et~al\mbox{.}(2025)]%
        {moura2025nexardashcamcollisionprediction}
\bibfield{author}{\bibinfo{person}{Daniel~C. Moura}, \bibinfo{person}{Shizhan Zhu}, {and} \bibinfo{person}{Orly Zvitia}.} \bibinfo{year}{2025}\natexlab{}.
\newblock \bibinfo{title}{Nexar Dashcam Collision Prediction Dataset and Challenge}.
\newblock
\showeprint[arxiv]{2503.03848}~[cs.CV]
\urldef\tempurl%
\url{https://arxiv.org/abs/2503.03848}
\showURL{%
\tempurl}


\bibitem[Muller et~al\mbox{.}(2022)]%
        {Muller2022}
\bibfield{author}{\bibinfo{person}{Emily Muller}, \bibinfo{person}{Emily Gemmell}, \bibinfo{person}{Ishmam Choudhury}, \bibinfo{person}{Ricky Nathvani}, \bibinfo{person}{Antje~Barbara Metzler}, \bibinfo{person}{James Bennett}, \bibinfo{person}{Emily Denton}, \bibinfo{person}{Seth Flaxman}, {and} \bibinfo{person}{Majid Ezzati}.} \bibinfo{year}{2022}\natexlab{}.
\newblock \bibinfo{booktitle}{\emph{City-{{Wide Perceptions}} of {{Neighbourhood Quality}} Using {{Street View Images}}}}. Vol.~\bibinfo{volume}{1}.
\newblock \bibinfo{publisher}{Association for Computing Machinery}.
\newblock
\showeprint[arxiv]{2211.12139}


\bibitem[Naik et~al\mbox{.}(2014)]%
        {Naik2014}
\bibfield{author}{\bibinfo{person}{Nikhil Naik}, \bibinfo{person}{Jade Philipoom}, \bibinfo{person}{Ramesh Raskar}, {and} \bibinfo{person}{Cesar Hidalgo}.} \bibinfo{year}{2014}\natexlab{}.
\newblock \showarticletitle{Streetscore-Predicting the Perceived Safety of One Million Streetscapes}.
\newblock \bibinfo{journal}{\emph{IEEE Computer Society Conference on Computer Vision and Pattern Recognition Workshops}} \bibinfo{number}{January} (\bibinfo{year}{2014}), \bibinfo{pages}{793--799}.
\newblock
\showISBNx{9781479943098}
\showISSN{21607516}
\href{https://doi.org/10.1109/CVPRW.2014.121}{doi:\nolinkurl{10.1109/CVPRW.2014.121}}


\bibitem[Neuhold et~al\mbox{.}(2017a)]%
        {vistas}
\bibfield{author}{\bibinfo{person}{Gerhard Neuhold}, \bibinfo{person}{Tobias Ollmann}, \bibinfo{person}{Samuel~Rota Bulò}, {and} \bibinfo{person}{Peter Kontschieder}.} \bibinfo{year}{2017}\natexlab{a}.
\newblock \showarticletitle{The Mapillary Vistas Dataset for Semantic Understanding of Street Scenes}. In \bibinfo{booktitle}{\emph{2017 IEEE International Conference on Computer Vision (ICCV)}}. \bibinfo{pages}{5000--5009}.
\newblock
\href{https://doi.org/10.1109/ICCV.2017.534}{doi:\nolinkurl{10.1109/ICCV.2017.534}}


\bibitem[Neuhold et~al\mbox{.}(2017b)]%
        {Neuhold_2017_ICCV}
\bibfield{author}{\bibinfo{person}{Gerhard Neuhold}, \bibinfo{person}{Tobias Ollmann}, \bibinfo{person}{Samuel Rota~Bulo}, {and} \bibinfo{person}{Peter Kontschieder}.} \bibinfo{year}{2017}\natexlab{b}.
\newblock \showarticletitle{The Mapillary Vistas Dataset for Semantic Understanding of Street Scenes}. In \bibinfo{booktitle}{\emph{Proceedings of the IEEE International Conference on Computer Vision (ICCV)}}.
\newblock


\bibitem[{OpenStreetMap contributors}(2017)]%
        {OpenStreetMap}
\bibfield{author}{\bibinfo{person}{{OpenStreetMap contributors}}.} \bibinfo{year}{2017}\natexlab{}.
\newblock \bibinfo{title}{{Planet dump retrieved from https://planet.osm.org }}.
\newblock \bibinfo{howpublished}{\url{ https://www.openstreetmap.org }}.
\newblock


\bibitem[Ordonez and Berg(2014)]%
        {Ordonez2014}
\bibfield{author}{\bibinfo{person}{Vicente Ordonez} {and} \bibinfo{person}{Tamara~L. Berg}.} \bibinfo{year}{2014}\natexlab{}.
\newblock \showarticletitle{Learning High-Level Judgments of Urban Perception}.
\newblock \bibinfo{journal}{\emph{Lecture Notes in Computer Science (including subseries Lecture Notes in Artificial Intelligence and Lecture Notes in Bioinformatics)}} \bibinfo{volume}{8694 LNCS}, \bibinfo{number}{PART 6} (\bibinfo{year}{2014}), \bibinfo{pages}{494--510}.
\newblock
\showISBNx{9783319105987}
\showISSN{16113349}
\href{https://doi.org/10.1007/978-3-319-10599-4_32}{doi:\nolinkurl{10.1007/978-3-319-10599-4_32}}


\bibitem[{Perez-Cruz}(2008)]%
        {perez-kl2008}
\bibfield{author}{\bibinfo{person}{Fernando {Perez-Cruz}}.} \bibinfo{year}{2008}\natexlab{}.
\newblock \showarticletitle{Kullback-{{Leibler}} Divergence Estimation of Continuous Distributions}. In \bibinfo{booktitle}{\emph{2008 {{IEEE International Symposium}} on {{Information Theory}}}}. \bibinfo{pages}{1666--1670}.
\newblock
\showISSN{2157-8117}
\href{https://doi.org/10.1109/ISIT.2008.4595271}{doi:\nolinkurl{10.1109/ISIT.2008.4595271}}


\bibitem[Quinn and Alvarez~Le{\'o}n(2019)]%
        {quinnEverySingleStreet2019}
\bibfield{author}{\bibinfo{person}{Sterling Quinn} {and} \bibinfo{person}{Luis Alvarez~Le{\'o}n}.} \bibinfo{year}{2019}\natexlab{}.
\newblock \showarticletitle{Every Single Street? {{Rethinking}} Full Coverage across Street-Level Imagery Platforms}.
\newblock \bibinfo{journal}{\emph{Transactions in GIS}} \bibinfo{volume}{23}, \bibinfo{number}{6} (\bibinfo{year}{2019}), \bibinfo{pages}{1251--1272}.
\newblock
\showISSN{1467-9671}
\href{https://doi.org/10.1111/tgis.12571}{doi:\nolinkurl{10.1111/tgis.12571}}


\bibitem[Randhawa et~al\mbox{.}(2024)]%
        {randhawa2024pavedunpaveddeeplearning}
\bibfield{author}{\bibinfo{person}{Sukanya Randhawa}, \bibinfo{person}{Eren Aygun}, \bibinfo{person}{Guntaj Randhawa}, \bibinfo{person}{Benjamin Herfort}, \bibinfo{person}{Sven Lautenbach}, {and} \bibinfo{person}{Alexander Zipf}.} \bibinfo{year}{2024}\natexlab{}.
\newblock \bibinfo{title}{Paved or unpaved? A Deep Learning derived Road Surface Global Dataset from Mapillary Street-View Imagery}.
\newblock
\showeprint[arxiv]{2410.19874}~[cs.CV]
\urldef\tempurl%
\url{https://arxiv.org/abs/2410.19874}
\showURL{%
\tempurl}


\bibitem[Rzotkiewicz et~al\mbox{.}(2018)]%
        {rzotkiewiczSystematicReviewUse2018}
\bibfield{author}{\bibinfo{person}{Amanda Rzotkiewicz}, \bibinfo{person}{Amber~L. Pearson}, \bibinfo{person}{Benjamin~V. Dougherty}, \bibinfo{person}{Ashton Shortridge}, {and} \bibinfo{person}{Nick Wilson}.} \bibinfo{year}{2018}\natexlab{}.
\newblock \showarticletitle{Systematic Review of the Use of {{Google Street View}} in Health Research: {{Major}} Themes, Strengths, Weaknesses and Possibilities for Future Research}.
\newblock \bibinfo{journal}{\emph{Health \& Place}}  \bibinfo{volume}{52} (\bibinfo{date}{July} \bibinfo{year}{2018}), \bibinfo{pages}{240--246}.
\newblock
\showISSN{13538292}
\href{https://doi.org/10.1016/j.healthplace.2018.07.001}{doi:\nolinkurl{10.1016/j.healthplace.2018.07.001}}


\bibitem[Sakurada(2018)]%
        {Sakurada2018}
\bibfield{author}{\bibinfo{person}{Ken Sakurada}.} \bibinfo{year}{2018}\natexlab{}.
\newblock \showarticletitle{Weakly {{Supervised Silhouette-based Semantic Change Detection}}}.
\newblock  (\bibinfo{year}{2018}).
\newblock
\showeprint[arxiv]{1811.11985}


\bibitem[Seiferling et~al\mbox{.}(2017)]%
        {Seiferling2017}
\bibfield{author}{\bibinfo{person}{Ian Seiferling}, \bibinfo{person}{Nikhil Naik}, \bibinfo{person}{Carlo Ratti}, {and} \bibinfo{person}{Raph{\"a}el Proulx}.} \bibinfo{year}{2017}\natexlab{}.
\newblock \showarticletitle{Green Streets - {{Quantifying}} and Mapping Urban Trees with Street-Level Imagery and Computer Vision}.
\newblock \bibinfo{journal}{\emph{Landscape and Urban Planning}} \bibinfo{volume}{165}, \bibinfo{number}{July 2016} (\bibinfo{year}{2017}), \bibinfo{pages}{93--101}.
\newblock
\showISSN{01692046}
\href{https://doi.org/10.1016/j.landurbplan.2017.05.010}{doi:\nolinkurl{10.1016/j.landurbplan.2017.05.010}}


\bibitem[Seresinhe et~al\mbox{.}(2017)]%
        {Seresinhe2017}
\bibfield{author}{\bibinfo{person}{Chanuki~Illushka Seresinhe}, \bibinfo{person}{Tobias Preis}, {and} \bibinfo{person}{Helen~Susannah Moat}.} \bibinfo{year}{2017}\natexlab{}.
\newblock \showarticletitle{Using Deep Learning to Quantify the Beauty of Outdoor Places}.
\newblock \bibinfo{journal}{\emph{Royal Society Open Science}} \bibinfo{volume}{4}, \bibinfo{number}{7} (\bibinfo{year}{2017}).
\newblock
\showISBNx{0000000191}
\showISSN{20545703}
\href{https://doi.org/10.1098/rsos.170170}{doi:\nolinkurl{10.1098/rsos.170170}}


\bibitem[Smith et~al\mbox{.}(2021)]%
        {gsvbronx}
\bibfield{author}{\bibinfo{person}{Cara~M. Smith}, \bibinfo{person}{Joel~D. Kaufman}, {and} \bibinfo{person}{Stephen~J. Mooney}.} \bibinfo{year}{2021}\natexlab{}.
\newblock \showarticletitle{Google street view image availability in the Bronx and San Diego, 2007--2020: Understanding potential biases in virtual audits of urban built environments}.
\newblock \bibinfo{journal}{\emph{Health \& Place}}  \bibinfo{volume}{72} (\bibinfo{year}{2021}), \bibinfo{pages}{102701}.
\newblock
\showISSN{1353-8292}
\href{https://doi.org/10.1016/j.healthplace.2021.102701}{doi:\nolinkurl{10.1016/j.healthplace.2021.102701}}


\bibitem[Suel et~al\mbox{.}(2019)]%
        {Suel2019}
\bibfield{author}{\bibinfo{person}{Esra Suel}, \bibinfo{person}{John~W. Polak}, \bibinfo{person}{James~E. Bennett}, {and} \bibinfo{person}{Majid Ezzati}.} \bibinfo{year}{2019}\natexlab{}.
\newblock \showarticletitle{Measuring Social, Environmental and Health Inequalities Using Deep Learning and Street Imagery}.
\newblock \bibinfo{journal}{\emph{Scientific Reports}} \bibinfo{volume}{9}, \bibinfo{number}{1} (\bibinfo{year}{2019}), \bibinfo{pages}{1--10}.
\newblock
\showISBNx{4159801942036}
\showISSN{20452322}
\href{https://doi.org/10.1038/s41598-019-42036-w}{doi:\nolinkurl{10.1038/s41598-019-42036-w}}


\bibitem[Sukel et~al\mbox{.}(2020)]%
        {Sukel2020}
\bibfield{author}{\bibinfo{person}{Maarten Sukel}, \bibinfo{person}{Stevan Rudinac}, {and} \bibinfo{person}{Marcel Worring}.} \bibinfo{year}{2020}\natexlab{}.
\newblock \showarticletitle{Urban Object Detection Kit: {{A}} System for Collection and Analysis of Street-Level Imagery}.
\newblock \bibinfo{journal}{\emph{ICMR 2020 - Proceedings of the 2020 International Conference on Multimedia Retrieval}} (\bibinfo{year}{2020}), \bibinfo{pages}{509--516}.
\newblock
\showISBNx{9781450370875}
\href{https://doi.org/10.1145/3372278.3390708}{doi:\nolinkurl{10.1145/3372278.3390708}}


\bibitem[Torii et~al\mbox{.}({[n.\,d.]})]%
        {Torii}
\bibfield{author}{\bibinfo{person}{Akihiko Torii}, \bibinfo{person}{Relja Arandjelovi}, \bibinfo{person}{Sivic Masatoshi}, {and} \bibinfo{person}{Okutomi Tomas}.} \bibinfo{year}{[n.\,d.]}\natexlab{}.
\newblock \showarticletitle{Torii-{{CVPR-2015-final}}}.
\newblock  (\bibinfo{year}{[n.\,d.]}).
\newblock


\bibitem[Torii et~al\mbox{.}(2013)]%
        {Torii2013}
\bibfield{author}{\bibinfo{person}{Akihiko Torii}, \bibinfo{person}{Josef Sivic}, \bibinfo{person}{Toma Pajdla}, {and} \bibinfo{person}{Masatoshi Okutomi}.} \bibinfo{year}{2013}\natexlab{}.
\newblock \showarticletitle{Visual Place Recognition with Repetitive Structures}.
\newblock \bibinfo{journal}{\emph{Proceedings of the IEEE Computer Society Conference on Computer Vision and Pattern Recognition}} (\bibinfo{year}{2013}), \bibinfo{pages}{883--890}.
\newblock
\showISSN{10636919}
\href{https://doi.org/10.1109/CVPR.2013.119}{doi:\nolinkurl{10.1109/CVPR.2013.119}}


\bibitem[Umar et~al\mbox{.}(2023)]%
        {umarPotentialGoogleStreet2023}
\bibfield{author}{\bibinfo{person}{Farouk Umar}, \bibinfo{person}{Josephine Amoah}, \bibinfo{person}{Moses Asamoah}, \bibinfo{person}{Mawuli Dzodzomenyo}, \bibinfo{person}{Chidinma Igwenagu}, \bibinfo{person}{Lorna-Grace Okotto}, \bibinfo{person}{Joseph {Okotto-Okotto}}, \bibinfo{person}{Pete Shaw}, {and} \bibinfo{person}{Jim Wright}.} \bibinfo{year}{2023}\natexlab{}.
\newblock \showarticletitle{On the Potential of {{Google Street View}} for Environmental Waste Quantification in Urban {{Africa}}: {{An}} Assessment of Bias in Spatial Coverage}.
\newblock \bibinfo{journal}{\emph{Sustainable Environment}} \bibinfo{volume}{9}, \bibinfo{number}{1} (\bibinfo{date}{Dec.} \bibinfo{year}{2023}), \bibinfo{pages}{2251799}.
\newblock
\showISSN{null}
\href{https://doi.org/10.1080/27658511.2023.2251799}{doi:\nolinkurl{10.1080/27658511.2023.2251799}}


\bibitem[Vaserstein(1969)]%
        {wasserstein}
\bibfield{author}{\bibinfo{person}{N. Vaserstein, L.}} \bibinfo{year}{1969}\natexlab{}.
\newblock \showarticletitle{Markov Processes over Denumerable Products of Spaces, Describing Large Systems of Automata}.
\newblock \bibinfo{journal}{\emph{Probl. Peredachi Inf.}} (\bibinfo{year}{1969}), \bibinfo{pages}{64--72}.
\newblock
\urldef\tempurl%
\url{http://mi.mathnet.ru/ppi1811}
\showURL{%
\tempurl}


\bibitem[Warburg et~al\mbox{.}(2020)]%
        {Warburg}
\bibfield{author}{\bibinfo{person}{Frederik Warburg}, \bibinfo{person}{Søren Hauberg}, \bibinfo{person}{Manuel López-Antequera}, \bibinfo{person}{Pau Gargallo}, \bibinfo{person}{Yubin Kuang}, {and} \bibinfo{person}{Javier Civera}.} \bibinfo{year}{2020}\natexlab{}.
\newblock \showarticletitle{Mapillary Street-Level Sequences: A Dataset for Lifelong Place Recognition}. In \bibinfo{booktitle}{\emph{2020 IEEE/CVF Conference on Computer Vision and Pattern Recognition (CVPR)}}. \bibinfo{pages}{2623--2632}.
\newblock
\href{https://doi.org/10.1109/CVPR42600.2020.00270}{doi:\nolinkurl{10.1109/CVPR42600.2020.00270}}


\bibitem[Weyand et~al\mbox{.}(2020)]%
        {GoogleLandmarksV2}
\bibfield{author}{\bibinfo{person}{Tobias Weyand}, \bibinfo{person}{Andr{\'{e}} Ara{\'{u}}jo}, \bibinfo{person}{Bingyi Cao}, {and} \bibinfo{person}{Jack Sim}.} \bibinfo{year}{2020}\natexlab{}.
\newblock \showarticletitle{Google Landmarks Dataset v2 - {A} Large-Scale Benchmark for Instance-Level Recognition and Retrieval}.
\newblock \bibinfo{journal}{\emph{CoRR}}  \bibinfo{volume}{abs/2004.01804} (\bibinfo{year}{2020}).
\newblock
\showeprint[arXiv]{2004.01804}
\urldef\tempurl%
\url{https://arxiv.org/abs/2004.01804}
\showURL{%
\tempurl}


\end{thebibliography}
\newpage
\phantom{hey}

\clearpage 
\appendix
\section{Appendix}
\subsection{Case Study Questionnaire}
\begin{table}[h]
    \begin{tabular}{p{13cm}}
        \hline
         \textbf{Questions}  \\
        \hline

        General Questions \\
         What does your company do?
         What is your role within the company?  \\
         What is the task your company has been set regarding the collecting of street view images for the Amsterdam Panorama Database?
           \\
        Could you guide us through the process from getting the objective from the municipality to collection by mapping out phases - and specific actions in each phase?
        What does the process look like ?
         \\\\
        I. Collection process  \\
         How do you determine which route the car drives?  \\
         How do you determine \textit{when} the car drives?  \\
         In what months do you drive?  \\
         At what time of day do you drive?  \\
         How much is captured in a single drive?  \\
         Do you drive multiple times per day?  \\
         Is a neighbourhood always captured in a single drive?  \\
         How long do you take to capture how much?  \\
         When capturing a neighbourhood in multiple drive, do you vary the entrance roads?  \\
         How many different drivers are there? 
         Are drivers allowed freedom in the way they drive? 
         \\\\
        
        II. Goals and requirements  \\
         What is the goal of collection?  \\
         Is there a document outlining the specifics of the goal or the details regarding the way the collection process should come about?  \\
         If so, what does this specify regarding the collection process?  \\
        Do you have a surplus of images? Are images thrown away before sending them to the municipality?  \\
         
         \\
        III. Technical questions  \\
        What are the camera specifications  \\
        Does the camera take viewpoint images and stitch them together?  \\
        Are the images stored as panoramas only? Are the viewpoint images stored also?
        How fast does the camera turn to make a single panorama?  \\
        How fast does the car drive during the collection process?  \\
        At what intervals are panoramas captured? Is this a temporal or spatial interval?  \\

    \bottomrule
    \end{tabular}
    \caption{Questions used for the semi-structured interviews.}
    \label{interviewquestions}
\end{table}

\clearpage
\subsection{Distribution Ranking}
\begin{table}[!hb]
    \centering
    \scriptsize
    \begin{tabular}{llrrrr}
    \toprule
    City & Provider & EMD & KL & EMD-ranking & KL-ranking \\
    \midrule
    Kiev & GSV & 0.000021 & 7.545364 & 13 & 1 \\
    Almaty & GSV & 0.000028 & 9.215246 & 15 & 2 \\
    Kiev & MLY & 0.000146 & 9.586454 & 21 & 3 \\
    Los Angeles & GSV & 0.002705 & 10.339047 & 46 & 4 \\
    Auckland & GSV & 0.001626 & 10.347198 & 43 & 5 \\
    Reykjavik & GSV & 0.000009 & 10.419424 & 8 & 6 \\
    Sydney & GSV & 0.001841 & 10.473597 & 44 & 7 \\
    Pittsburgh & GSV & 0.000007 & 12.142508 & 6 & 8 \\
    Istanbul & GSV & 0.003657 & 12.155309 & 50 & 9 \\
    Lagos & GSV & 0.003334 & 13.206205 & 49 & 10 \\
    Nairobi & GSV & 0.000192 & 13.255246 & 23 & 11 \\
    Dhaka & GSV & 0.001127 & 13.319114 & 38 & 12 \\
    Johannesburg & GSV & 0.001216 & 13.496110 & 39 & 13 \\
    Bangkok & GSV & 0.000100 & 13.571742 & 19 & 14 \\
    Almaty & MLY & 0.000307 & 13.664120 & 26 & 15 \\
    London & GSV & 0.000022 & 13.704638 & 14 & 16 \\
    Reykjavik & MLY & 0.000510 & 14.376512 & 28 & 17 \\
    Los Angeles & MLY & 0.003057 & 14.547335 & 47 & 18 \\
    Mexico City & GSV & 0.000582 & 14.574151 & 32 & 19 \\
    Vancouver & GSV & 0.000004 & 14.581596 & 3 & 20 \\
    Dubai & MLY & 0.000865 & 14.732709 & 35 & 21 \\
    Auckland & MLY & 0.005057 & 14.884889 & 54 & 22 \\
    Dubai & GSV & 0.000550 & 14.970169 & 31 & 23 \\
    Rio de Janeiro & GSV & 0.000474 & 15.354438 & 27 & 24 \\
    Istanbul & MLY & 0.009901 & 15.698075 & 56 & 25 \\
    Pittsburgh & MLY & 0.000010 & 15.785522 & 10 & 26 \\
    Sydney & MLY & 0.006922 & 15.791619 & 55 & 27 \\
    Nairobi & MLY & 0.003071 & 15.918315 & 48 & 28 \\
    Seoul & GSV & 0.000007 & 16.053656 & 7 & 29 \\
    New Delhi & GSV & 0.000009 & 16.265394 & 9 & 30 \\
    Amsterdam & MLY & 0.000006 & 16.379862 & 4 & 31 \\
    Amsterdam & GSV & 0.000004 & 16.428343 & 2 & 32 \\
    Dhaka & MLY & 0.001424 & 16.734468 & 40 & 33 \\
    Johannesburg & MLY & 0.002221 & 17.410635 & 45 & 34 \\
    New Delhi & MLY & 0.000185 & 17.471664 & 22 & 35 \\
    Lima & GSV & 0.000900 & 17.526670 & 36 & 36 \\
    Tokyo & GSV & 0.000525 & 17.666317 & 30 & 37 \\
    Paris & GSV & 0.000006 & 18.060394 & 5 & 38 \\
    Buenos Aires & GSV & 0.000001 & 18.468925 & 1 & 39 \\
    Vancouver & MLY & 0.000112 & 18.490164 & 20 & 40 \\
    Bangkok & MLY & 0.001014 & 18.511465 & 37 & 41 \\
    Rio de Janeiro & MLY & 0.003918 & 18.580067 & 52 & 42 \\
    Accra & GSV & 0.000019 & 18.687561 & 12 & 43 \\
    Lagos & MLY & 0.004466 & 18.731262 & 53 & 44 \\
    Accra & MLY & 0.000092 & 18.788965 & 18 & 45 \\
    London & MLY & 0.000850 & 18.817217 & 34 & 46 \\
    Mexico City & MLY & 0.001472 & 19.359091 & 41 & 47 \\
    Dakar & GSV & 0.000265 & 20.407593 & 25 & 48 \\
    Paris & MLY & 0.000058 & 20.848209 & 17 & 49 \\
    Buenos Aires & MLY & 0.000011 & 21.225250 & 11 & 50 \\
    Lima & MLY & 0.001603 & 21.332228 & 42 & 51 \\
    Seoul & MLY & 0.000522 & 21.878321 & 29 & 52 \\
    Tokyo & MLY & 0.000788 & 21.963434 & 33 & 53 \\
    Singapore & GSV & 0.000032 & 23.527073 & 16 & 54 \\
    Dakar & MLY & 0.003811 & 25.120543 & 51 & 55 \\
    Singapore & MLY & 0.000255 & 25.684063 & 24 & 56 \\
    \bottomrule
    \end{tabular}

    \caption{Results of the evaluation of the distance between Uniform Road Coverage and Real coverage on driveable roads for our selected cities. Scores are ranked individually. The table is sorted based on KL Rank.}
    \label{tab:FullRank-Table}
\end{table}

\begin{table}[t]
    \centering
    \scriptsize
    \begin{tabular}{llrrrr}
    \toprule
    City & Provider & EMD & KL & EMD-ranking & KL-ranking \\
    \midrule
    Dhaka & GSV & 0.001127 & 14.039197 & 37 & 1 \\
    Istanbul & GSV & 0.003657 & 14.306140 & 50 & 2 \\
    Lagos & GSV & 0.003334 & 14.403610 & 49 & 3 \\
    Almaty & GSV & 0.000028 & 15.810487 & 15 & 4 \\
    Nairobi & GSV & 0.000192 & 17.045719 & 22 & 5 \\
    Istanbul & MLY & 0.009901 & 17.098923 & 56 & 6 \\
    Bangkok & GSV & 0.000100 & 17.112480 & 18 & 7 \\
    Johannesburg & GSV & 0.001216 & 17.292749 & 38 & 8 \\
    Los Angeles & GSV & 0.002705 & 17.448380 & 46 & 9 \\
    Almaty & MLY & 0.000307 & 18.524441 & 24 & 10 \\
    Rio de Janeiro & GSV & 0.000474 & 18.883480 & 25 & 11 \\
    Nairobi & MLY & 0.003071 & 18.893942 & 48 & 12 \\
    Mexico City & GSV & 0.000582 & 19.146158 & 30 & 13 \\
    Dubai & MLY & 0.000865 & 19.170399 & 34 & 14 \\
    Dubai & GSV & 0.000550 & 19.233101 & 29 & 15 \\
    New Delhi & GSV & 0.000009 & 19.305271 & 9 & 16 \\
    Lagos & MLY & 0.004466 & 19.450876 & 53 & 17 \\
    New Delhi & MLY & 0.000185 & 19.880428 & 21 & 18 \\
    Dhaka & MLY & 0.001424 & 20.118916 & 39 & 19 \\
    Johannesburg & MLY & 0.002221 & 20.536474 & 44 & 20 \\
    Dakar & GSV & 0.000265 & 20.548519 & 23 & 21 \\
    Los Angeles & MLY & 0.003057 & 20.712873 & 47 & 22 \\
    Sydney & GSV & 0.001841 & 20.749775 & 43 & 23 \\
    Seoul & GSV & 0.000007 & 21.138929 & 7 & 24 \\
    Bangkok & MLY & 0.001014 & 21.311998 & 36 & 25 \\
    Rio de Janeiro & MLY & 0.003918 & 21.504551 & 52 & 26 \\
    Accra & MLY & 0.000092 & 21.859734 & 17 & 27 \\
    Accra & GSV & 0.000019 & 22.090065 & 12 & 28 \\
    Auckland & GSV & 0.001626 & 22.559696 & 42 & 29 \\
    Mexico City & MLY & 0.001472 & 23.071579 & 40 & 30 \\
    Buenos Aires & GSV & 0.000001 & 23.211132 & 1 & 31 \\
    Lima & GSV & 0.000900 & 24.182022 & 35 & 32 \\
    Kiev & GSV & 0.000021 & 24.360056 & 13 & 33 \\
    Sydney & MLY & 0.006922 & 24.388742 & 55 & 34 \\
    Auckland & MLY & 0.005057 & 24.788319 & 54 & 35 \\
    Dakar & MLY & 0.003811 & 24.803247 & 51 & 36 \\
    Seoul & MLY & 0.000522 & 25.452660 & 27 & 37 \\
    Buenos Aires & MLY & 0.000011 & 25.566809 & 11 & 38 \\
    Tokyo & GSV & 0.000525 & 26.212103 & 28 & 39 \\
    Amsterdam & MLY & 0.000006 & 27.371927 & 4 & 40 \\
    Lima & MLY & 0.001603 & 27.461069 & 41 & 41 \\
    Reykjavik & GSV & 0.000009 & 27.902699 & 8 & 42 \\
    Pittsburgh & GSV & 0.000007 & 28.119865 & 6 & 43 \\
    Amsterdam & GSV & 0.000004 & 28.179003 & 2 & 44 \\
    London & GSV & 0.000022 & 29.124451 & 14 & 45 \\
    Tokyo & MLY & 0.000788 & 29.148968 & 31 & 46 \\
    Pittsburgh & MLY & 0.000010 & 30.538847 & 10 & 47 \\
    Reykjavik & MLY & 0.000510 & 30.882397 & 26 & 48 \\
    Kiev & MLY & 0.000146 & 35.098461 & 20 & 49 \\
    Singapore  & GSV & 0.000846 & 35.171696 & 32 & 50 \\
    Vancouver & GSV & 0.000004 & 36.003235 & 3 & 51 \\
    Singapore  & MLY & 0.002372 & 36.650494 & 45 & 52 \\
    Vancouver & MLY & 0.000112 & 38.404789 & 19 & 53 \\
    London & MLY & 0.000850 & 39.920349 & 33 & 54 \\
    Paris & GSV & 0.000006 & 62.474991 & 5 & 55 \\
    Paris & MLY & 0.000058 & 63.422729 & 16 & 56 \\
    \bottomrule
    \end{tabular}
    
    \caption{Results of the evaluation of the distance between Uniform Road Coverage and Real coverage on all publicly accessible roads for our selected cities. Scores are ranked individually. The table is sorted based on KL Rank.}
    \label{tab:FullRank-Table_public}
\end{table}

\clearpage
\onecolumn 
\subsection{Correlation Plots}

\begin{figure}[!h]
    \includegraphics[width=.9\textwidth]
{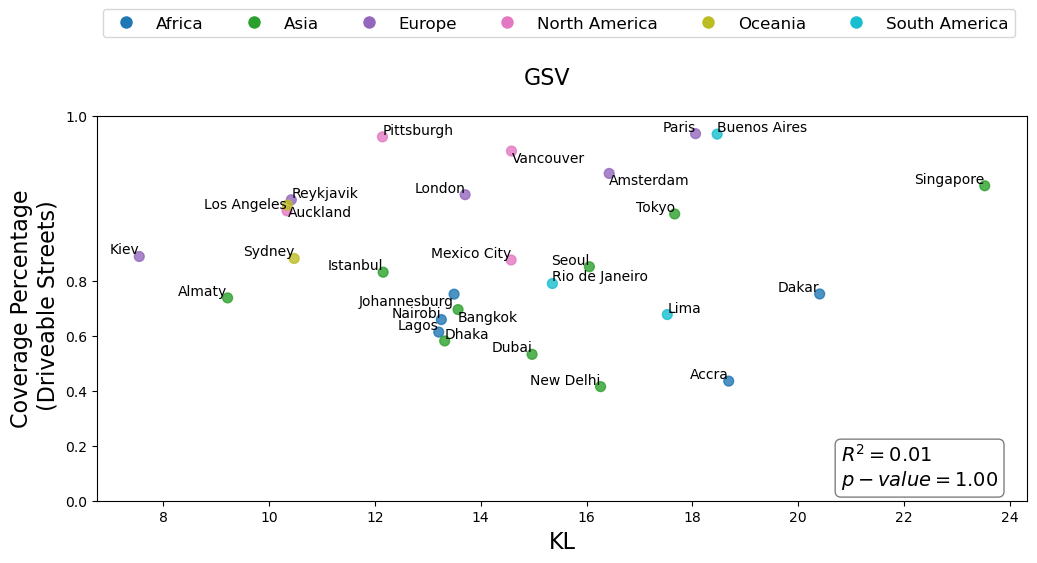}
    \caption{Coverage percentages plotted against the KL Divergence for Google Street View for all driveable streets.}
    \label{fig:app_GSV_Drive_KL_firsttt}
\end{figure}

\begin{figure}
    \includegraphics[width=0.9\linewidth]{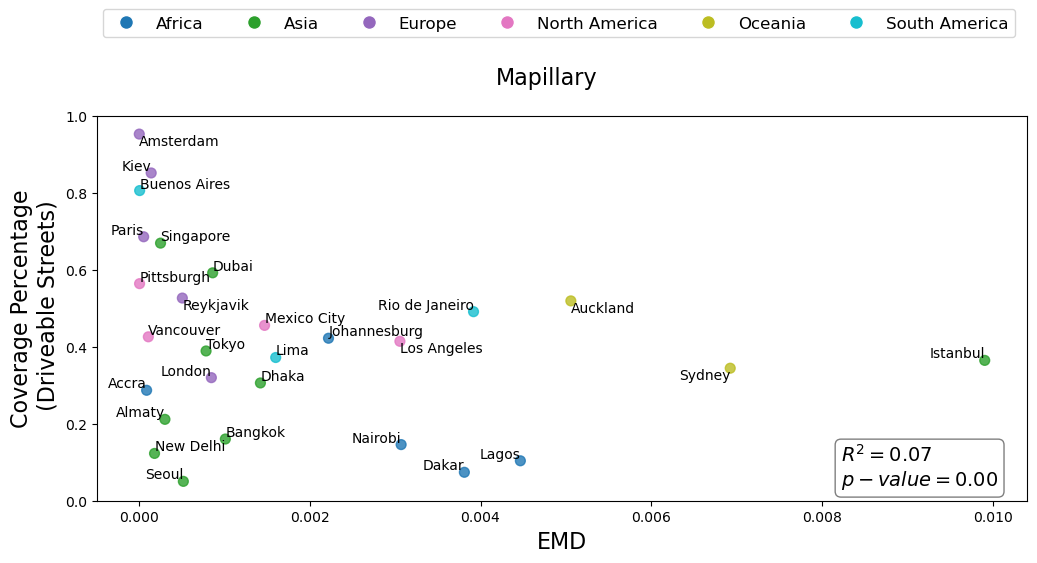}
\caption{Coverage percentages plotted against the EMD for Mapillary for all driveable streets.}
        \label{fig:app_MLY_Drive_EMD}
\end{figure}
\begin{figure}
    \includegraphics[width=0.9\linewidth]{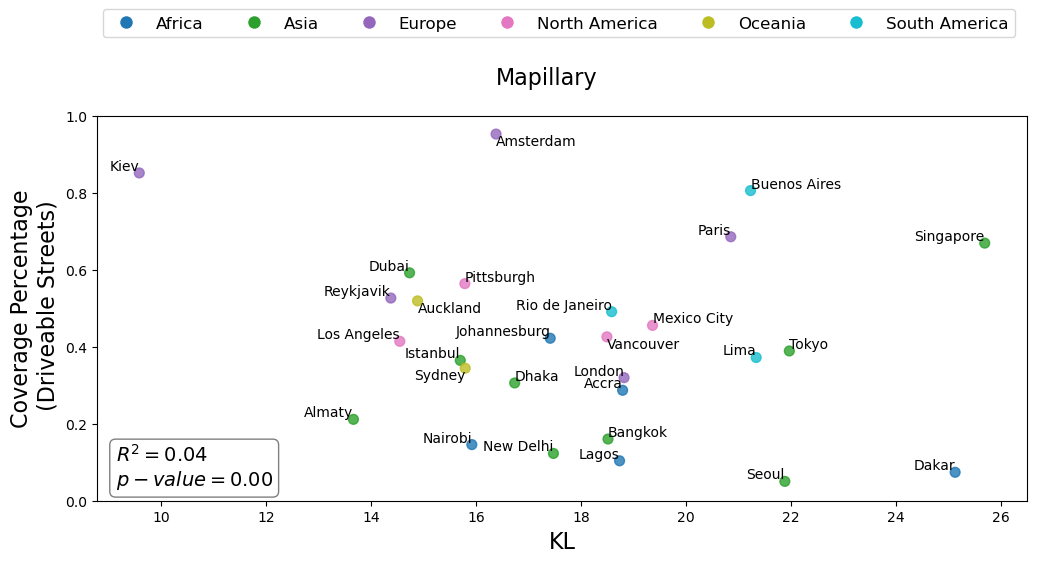}

\caption{Coverage percentages plotted against the KL Divergence for Mapillary for all driveable streets.}
    \label{fig:app_MLY_Drive_KL}
\end{figure}

\begin{figure}
    \includegraphics[width=0.9\linewidth]{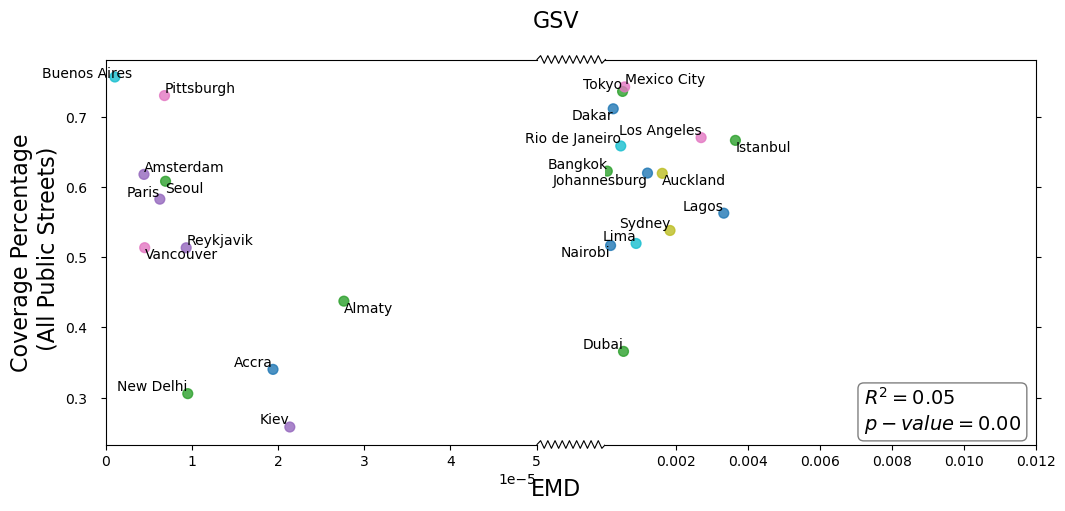}

        \caption{Coverage percentages plotted against the EMD for Google Street View for all publicly accessible streets.}
            \label{fig:app_GSV_Pub_EMD}
\end{figure}

\begin{figure}
    \includegraphics[width=0.9\linewidth]{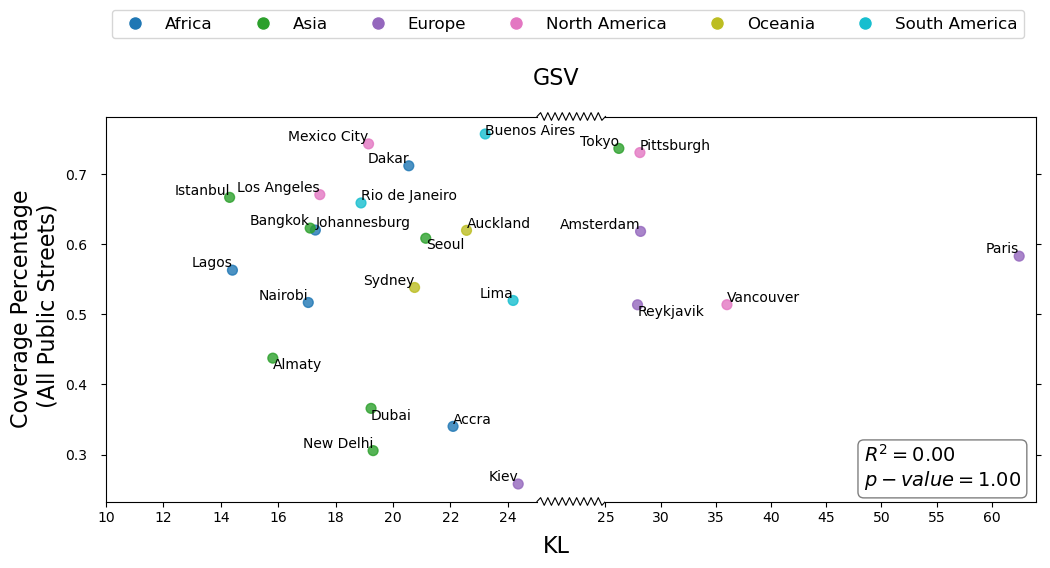}

        \caption{Coverage percentages plotted against the KL Divergence for Google Street View for publicly accessible streets.}
            \label{fig:app_GSV_Pub_KL}
\end{figure}


\begin{figure}
    \includegraphics[width=0.9\linewidth]{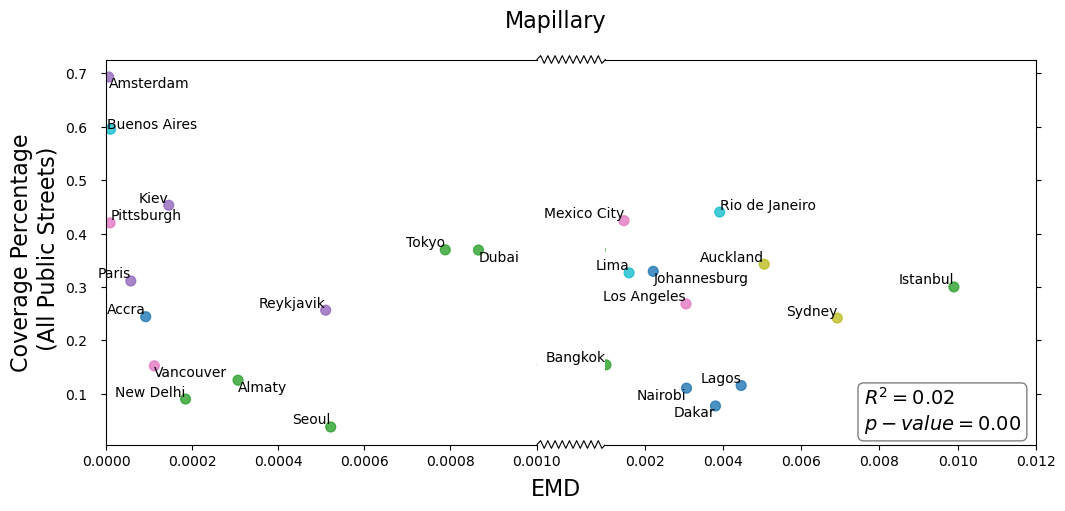}

        \caption{Coverage percentages plotted against the EMD for Mapillary for all publicly accessible streets.}
            \label{fig:app_MLY_Pub_EMD}
\end{figure}

\begin{figure}
    \includegraphics[width=0.9\linewidth]{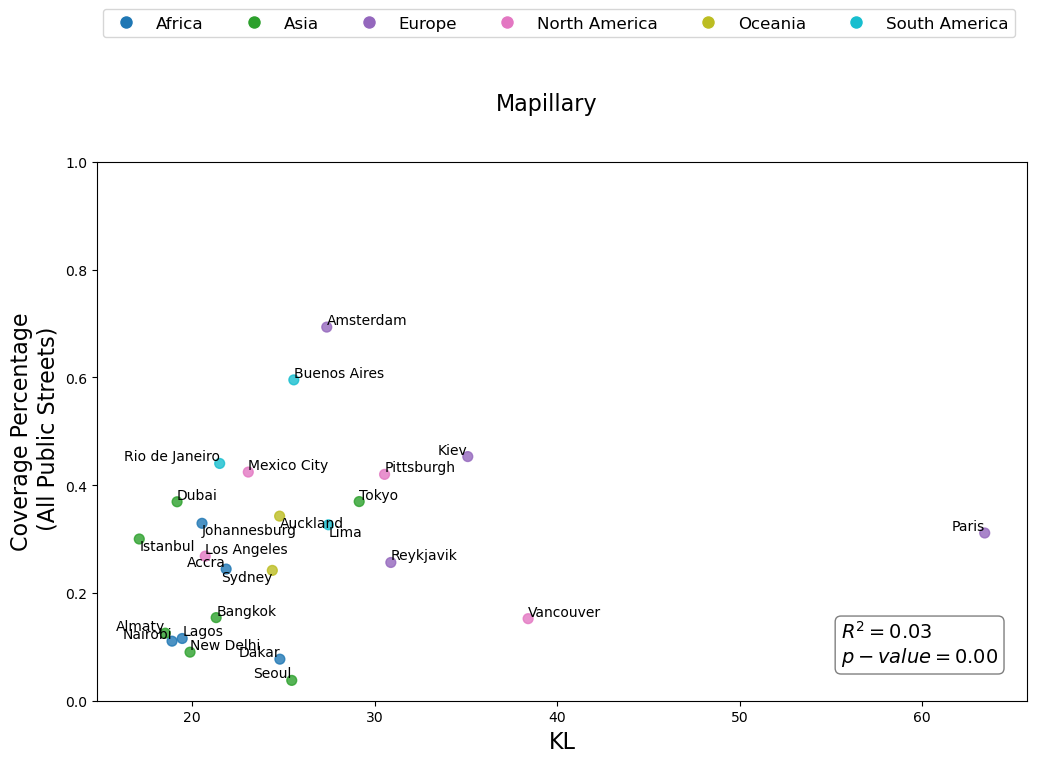}

    \caption{Coverage percentages plotted against the KL Divergence for Mapillary for publicly accessible streets.}
        \label{fig:app_MLY_Pub_KL_lasttt}
\end{figure}

\clearpage
\subsection{Significance scores}

\begin{table}[!hb]
    \centering
    \begin{tabular}{llllllllll}
    \toprule
     &  & \multicolumn{2}{c}{Wilks' Lambda} & \multicolumn{2}{c}{Pillai's Trace} & \multicolumn{2}{c}{Hotelling-Lawley Trace} & \multicolumn{2}{c}{Roy's Greatest Root} \\
     & city & value & p-value & value & p-value & value & p-value & value & p-value \\
    \midrule
    0 & Dakar & 0.988 & 0.000 & 0.012 & 0.000 & 0.013 & 0.000 & 0.013 & 0.000 \\
    1 & Dubai & 0.985 & 0.000 & 0.015 & 0.000 & 0.015 & 0.000 & 0.015 & 0.000 \\
    2 & Johannesburg & 0.980 & 0.000 & 0.020 & 0.000 & 0.020 & 0.000 & 0.020 & 0.000 \\
    3 & Auckland & 0.995 & 0.000 & 0.005 & 0.000 & 0.005 & 0.000 & 0.005 & 0.000 \\
    4 & Lima & 0.950 & 0.000 & 0.050 & 0.000 & 0.053 & 0.000 & 0.053 & 0.000 \\
    5 & LosAngeles & 0.988 & 0.000 & 0.012 & 0.000 & 0.012 & 0.000 & 0.012 & 0.000 \\
    6 & Istanbul & 0.981 & 0.000 & 0.019 & 0.000 & 0.019 & 0.000 & 0.019 & 0.000 \\
    7 & Amsterdam & 0.999 & 0.000 & 0.001 & 0.000 & 0.001 & 0.000 & 0.001 & 0.000 \\
    8 & Kiev & 0.999 & 0.000 & 0.001 & 0.000 & 0.001 & 0.000 & 0.001 & 0.000 \\
    9 & Tokyo & 0.994 & 0.000 & 0.006 & 0.000 & 0.006 & 0.000 & 0.006 & 0.000 \\
    10 & GreaterSydney & 0.990 & 0.000 & 0.010 & 0.000 & 0.010 & 0.000 & 0.010 & 0.000 \\
    11 & Pittsburgh & 0.998 & 0.000 & 0.002 & 0.000 & 0.002 & 0.000 & 0.002 & 0.000 \\
    12 & MexicoCity & 0.970 & 0.000 & 0.030 & 0.000 & 0.030 & 0.000 & 0.030 & 0.000 \\
    13 & London & 1.000 & 0.000 & 0.000 & 0.000 & 0.000 & 0.000 & 0.000 & 0.000 \\
    14 & Dhaka & 0.976 & 0.000 & 0.024 & 0.000 & 0.024 & 0.000 & 0.024 & 0.000 \\
    15 & Lagos & 0.970 & 0.000 & 0.030 & 0.000 & 0.031 & 0.000 & 0.031 & 0.000 \\
    16 & Singapore & 0.999 & 0.000 & 0.001 & 0.000 & 0.001 & 0.000 & 0.001 & 0.000 \\
    17 & Seoul & 1.000 & 0.001 & 0.000 & 0.001 & 0.000 & 0.001 & 0.000 & 0.001 \\
    18 & Paris & 0.999 & 0.000 & 0.001 & 0.000 & 0.001 & 0.000 & 0.001 & 0.000 \\
    19 & NewDelhi & 0.998 & 0.000 & 0.002 & 0.000 & 0.002 & 0.000 & 0.002 & 0.000 \\
    20 & Nairobi & 0.992 & 0.000 & 0.008 & 0.000 & 0.008 & 0.000 & 0.008 & 0.000 \\
    21 & Accra & 0.994 & 0.000 & 0.006 & 0.000 & 0.006 & 0.000 & 0.006 & 0.000 \\
    22 & BuenosAires & 1.000 & 0.000 & 0.000 & 0.000 & 0.000 & 0.000 & 0.000 & 0.000 \\
    23 & Reykjavik & 0.997 & 0.000 & 0.003 & 0.000 & 0.003 & 0.000 & 0.003 & 0.000 \\
    24 & Bangkok & 0.999 & 0.000 & 0.001 & 0.000 & 0.001 & 0.000 & 0.001 & 0.000 \\
    25 & Vancouver & 0.999 & 0.000 & 0.001 & 0.000 & 0.001 & 0.000 & 0.001 & 0.000 \\
    26 & Almaty & 0.997 & 0.000 & 0.003 & 0.000 & 0.003 & 0.000 & 0.003 & 0.000 \\
    27 & RioDeJaneiro & 0.990 & 0.000 & 0.010 & 0.000 & 0.010 & 0.000 & 0.010 & 0.000 \\
    \bottomrule
    \end{tabular}

    \caption{Scores for the multivariate analysis of variance (MANOVA) for determining whether the distribution of available images differs significantly from the distribution of all driveable streets for Google Street View.}
    \label{tab:sig-gsv-drive}
\end{table}

\begin{table}
    \centering
    \begin{tabular}{llllllllll}
    \toprule
     &  & \multicolumn{2}{c}{Wilks' Lambda} & \multicolumn{2}{c}{Pillai's Trace} & \multicolumn{2}{c}{Hotelling-Lawley Trace} & \multicolumn{2}{c}{Roy's Greatest Root} \\
     & city & value & p-value & value & p-value & value & p-value & value & p-value \\
    \midrule
    0 & Dakar & 0.990 & 0.000 & 0.010 & 0.000 & 0.010 & 0.000 & 0.010 & 0.000 \\
    1 & Dubai & 0.973 & 0.000 & 0.027 & 0.000 & 0.028 & 0.000 & 0.028 & 0.000 \\
    2 & Johannesburg & 0.990 & 0.000 & 0.010 & 0.000 & 0.010 & 0.000 & 0.010 & 0.000 \\
    3 & Auckland & 0.995 & 0.000 & 0.005 & 0.000 & 0.005 & 0.000 & 0.005 & 0.000 \\
    4 & Lima & 0.970 & 0.000 & 0.030 & 0.000 & 0.031 & 0.000 & 0.031 & 0.000 \\
    5 & LosAngeles & 0.989 & 0.000 & 0.011 & 0.000 & 0.011 & 0.000 & 0.011 & 0.000 \\
    6 & Istanbul & 0.959 & 0.000 & 0.041 & 0.000 & 0.043 & 0.000 & 0.043 & 0.000 \\
    7 & Amsterdam & 0.999 & 0.000 & 0.001 & 0.000 & 0.001 & 0.000 & 0.001 & 0.000 \\
    8 & Kiev & 0.997 & 0.000 & 0.003 & 0.000 & 0.003 & 0.000 & 0.003 & 0.000 \\
    9 & Tokyo & 0.991 & 0.000 & 0.009 & 0.000 & 0.009 & 0.000 & 0.009 & 0.000 \\
    10 & GreaterSydney & 0.986 & 0.000 & 0.014 & 0.000 & 0.014 & 0.000 & 0.014 & 0.000 \\
    11 & Pittsburgh & 0.996 & 0.000 & 0.004 & 0.000 & 0.004 & 0.000 & 0.004 & 0.000 \\
    12 & MexicoCity & 0.973 & 0.000 & 0.027 & 0.000 & 0.028 & 0.000 & 0.028 & 0.000 \\
    13 & London & 1.000 & 0.000 & 0.000 & 0.000 & 0.000 & 0.000 & 0.000 & 0.000 \\
    14 & Dhaka & 0.976 & 0.000 & 0.024 & 0.000 & 0.025 & 0.000 & 0.025 & 0.000 \\
    15 & Lagos & 0.966 & 0.000 & 0.034 & 0.000 & 0.035 & 0.000 & 0.035 & 0.000 \\
    16 & Singapore & 0.998 & 0.000 & 0.002 & 0.000 & 0.002 & 0.000 & 0.002 & 0.000 \\
    17 & Seoul & 1.000 & 0.006 & 0.000 & 0.006 & 0.000 & 0.006 & 0.000 & 0.006 \\
    18 & Paris & 0.996 & 0.000 & 0.004 & 0.000 & 0.004 & 0.000 & 0.004 & 0.000 \\
    19 & NewDelhi & 0.994 & 0.000 & 0.006 & 0.000 & 0.006 & 0.000 & 0.006 & 0.000 \\
    20 & Nairobi & 0.998 & 0.000 & 0.002 & 0.000 & 0.002 & 0.000 & 0.002 & 0.000 \\
    21 & Accra & 0.997 & 0.000 & 0.003 & 0.000 & 0.003 & 0.000 & 0.003 & 0.000 \\
    22 & BuenosAires & 0.998 & 0.000 & 0.002 & 0.000 & 0.002 & 0.000 & 0.002 & 0.000 \\
    23 & Reykjavik & 0.997 & 0.000 & 0.003 & 0.000 & 0.003 & 0.000 & 0.003 & 0.000 \\
    24 & Bangkok & 0.999 & 0.000 & 0.001 & 0.000 & 0.001 & 0.000 & 0.001 & 0.000 \\
    25 & Vancouver & 0.999 & 0.000 & 0.001 & 0.000 & 0.001 & 0.000 & 0.001 & 0.000 \\
    26 & Almaty & 0.975 & 0.000 & 0.025 & 0.000 & 0.025 & 0.000 & 0.025 & 0.000 \\
    27 & RioDeJaneiro & 0.995 & 0.000 & 0.005 & 0.000 & 0.005 & 0.000 & 0.005 & 0.000 \\
    \bottomrule
    \end{tabular}

    \caption{Scores for the multivariate analysis of variance (MANOVA) for determining whether the distribution of available images differs significantly from the distribution of all public streets for Google Street View.}
    \label{tab:sig-gsv-public}
\end{table}

\begin{table}
    \centering
    \begin{tabular}{llllllllll}
    \toprule
     &  & \multicolumn{2}{c}{Wilks' Lambda} & \multicolumn{2}{c}{Pillai's Trace} & \multicolumn{2}{c}{Hotelling-Lawley Trace} & \multicolumn{2}{c}{Roy's Greatest Root} \\
     & city & value & p-value & value & p-value & value & p-value & value & p-value \\
    \midrule
    0 & Dakar & 0.829 & 0.000 & 0.171 & 0.000 & 0.206 & 0.000 & 0.206 & 0.000 \\
    1 & Dubai & 0.987 & 0.000 & 0.013 & 0.000 & 0.014 & 0.000 & 0.014 & 0.000 \\
    2 & Johannesburg & 0.973 & 0.000 & 0.027 & 0.000 & 0.027 & 0.000 & 0.027 & 0.000 \\
    3 & Auckland & 0.991 & 0.000 & 0.009 & 0.000 & 0.010 & 0.000 & 0.010 & 0.000 \\
    4 & Lima & 0.920 & 0.000 & 0.080 & 0.000 & 0.087 & 0.000 & 0.087 & 0.000 \\
    5 & LosAngeles & 0.989 & 0.000 & 0.011 & 0.000 & 0.011 & 0.000 & 0.011 & 0.000 \\
    6 & Istanbul & 0.942 & 0.000 & 0.058 & 0.000 & 0.062 & 0.000 & 0.062 & 0.000 \\
    7 & Amsterdam & 0.998 & 0.000 & 0.002 & 0.000 & 0.002 & 0.000 & 0.002 & 0.000 \\
    8 & Kiev & 0.994 & 0.000 & 0.006 & 0.000 & 0.006 & 0.000 & 0.006 & 0.000 \\
    9 & Tokyo & 0.988 & 0.000 & 0.012 & 0.000 & 0.012 & 0.000 & 0.012 & 0.000 \\
    10 & GreaterSydney & 0.955 & 0.000 & 0.045 & 0.000 & 0.047 & 0.000 & 0.047 & 0.000 \\
    11 & Pittsburgh & 0.997 & 0.000 & 0.003 & 0.000 & 0.003 & 0.000 & 0.003 & 0.000 \\
    12 & MexicoCity & 0.930 & 0.000 & 0.070 & 0.000 & 0.076 & 0.000 & 0.076 & 0.000 \\
    13 & London & 0.997 & 0.000 & 0.003 & 0.000 & 0.003 & 0.000 & 0.003 & 0.000 \\
    14 & Dhaka & 0.982 & 0.000 & 0.018 & 0.000 & 0.018 & 0.000 & 0.018 & 0.000 \\
    15 & Lagos & 0.996 & 0.000 & 0.004 & 0.000 & 0.004 & 0.000 & 0.004 & 0.000 \\
    16 & Singapore & 0.975 & 0.000 & 0.025 & 0.000 & 0.025 & 0.000 & 0.025 & 0.000 \\
    17 & Seoul & 0.993 & 0.000 & 0.007 & 0.000 & 0.007 & 0.000 & 0.007 & 0.000 \\
    18 & Paris & 0.974 & 0.000 & 0.026 & 0.000 & 0.026 & 0.000 & 0.026 & 0.000 \\
    19 & NewDelhi & 0.957 & 0.000 & 0.043 & 0.000 & 0.044 & 0.000 & 0.044 & 0.000 \\
    20 & Nairobi & 0.818 & 0.000 & 0.182 & 0.000 & 0.222 & 0.000 & 0.222 & 0.000 \\
    21 & Accra & 0.983 & 0.000 & 0.017 & 0.000 & 0.017 & 0.000 & 0.017 & 0.000 \\
    22 & BuenosAires & 0.997 & 0.000 & 0.003 & 0.000 & 0.003 & 0.000 & 0.003 & 0.000 \\
    23 & Reykjavik & 0.941 & 0.000 & 0.059 & 0.000 & 0.063 & 0.000 & 0.063 & 0.000 \\
    24 & Bangkok & 0.998 & 0.000 & 0.002 & 0.000 & 0.002 & 0.000 & 0.002 & 0.000 \\
    25 & Vancouver & 0.944 & 0.000 & 0.056 & 0.000 & 0.059 & 0.000 & 0.059 & 0.000 \\
    26 & Almaty & 0.974 & 0.000 & 0.026 & 0.000 & 0.026 & 0.000 & 0.026 & 0.000 \\
    27 & RioDeJaneiro & 0.913 & 0.000 & 0.087 & 0.000 & 0.096 & 0.000 & 0.096 & 0.000 \\
    \bottomrule
    \end{tabular}
    \caption{Scores for the multivariate analysis of variance (MANOVA) for determining whether the distribution of available images differs significantly from the distribution of all driveable streets for Mapillary.}
    \label{tab:sig-mly-drive}
\end{table}

\begin{table}
    \centering
    \begin{tabular}{llllllllll}
    \toprule
     &  & \multicolumn{2}{c}{Wilks' Lambda} & \multicolumn{2}{c}{Pillai's Trace} & \multicolumn{2}{c}{Hotelling-Lawley Trace} & \multicolumn{2}{c}{Roy's Greatest Root} \\
     & city & value & p-value & value & p-value & value & p-value & value & p-value \\
    \midrule
    0 & Dakar & 0.843 & 0.000 & 0.157 & 0.000 & 0.186 & 0.000 & 0.186 & 0.000 \\
    1 & Dubai & 0.975 & 0.000 & 0.025 & 0.000 & 0.026 & 0.000 & 0.026 & 0.000 \\
    2 & Johannesburg & 0.986 & 0.000 & 0.014 & 0.000 & 0.014 & 0.000 & 0.014 & 0.000 \\
    3 & Auckland & 0.992 & 0.000 & 0.008 & 0.000 & 0.008 & 0.000 & 0.008 & 0.000 \\
    4 & Lima & 0.940 & 0.000 & 0.060 & 0.000 & 0.063 & 0.000 & 0.063 & 0.000 \\
    5 & LosAngeles & 0.990 & 0.000 & 0.010 & 0.000 & 0.010 & 0.000 & 0.010 & 0.000 \\
    6 & Istanbul & 0.912 & 0.000 & 0.088 & 0.000 & 0.096 & 0.000 & 0.096 & 0.000 \\
    7 & Amsterdam & 0.999 & 0.000 & 0.001 & 0.000 & 0.001 & 0.000 & 0.001 & 0.000 \\
    8 & Kiev & 0.985 & 0.000 & 0.015 & 0.000 & 0.015 & 0.000 & 0.015 & 0.000 \\
    9 & Tokyo & 0.982 & 0.000 & 0.018 & 0.000 & 0.019 & 0.000 & 0.019 & 0.000 \\
    10 & GreaterSydney & 0.953 & 0.000 & 0.047 & 0.000 & 0.049 & 0.000 & 0.049 & 0.000 \\
    11 & Pittsburgh & 0.996 & 0.000 & 0.004 & 0.000 & 0.004 & 0.000 & 0.004 & 0.000 \\
    12 & MexicoCity & 0.936 & 0.000 & 0.064 & 0.000 & 0.068 & 0.000 & 0.068 & 0.000 \\
    13 & London & 0.996 & 0.000 & 0.004 & 0.000 & 0.004 & 0.000 & 0.004 & 0.000 \\
    14 & Dhaka & 0.983 & 0.000 & 0.017 & 0.000 & 0.018 & 0.000 & 0.018 & 0.000 \\
    15 & Lagos & 0.991 & 0.000 & 0.009 & 0.000 & 0.010 & 0.000 & 0.010 & 0.000 \\
    16 & Singapore & 0.969 & 0.000 & 0.031 & 0.000 & 0.032 & 0.000 & 0.032 & 0.000 \\
    17 & Seoul & 0.992 & 0.000 & 0.008 & 0.000 & 0.009 & 0.000 & 0.009 & 0.000 \\
    18 & Paris & 0.988 & 0.000 & 0.012 & 0.000 & 0.012 & 0.000 & 0.012 & 0.000 \\
    19 & NewDelhi & 0.945 & 0.000 & 0.055 & 0.000 & 0.058 & 0.000 & 0.058 & 0.000 \\
    20 & Nairobi & 0.863 & 0.000 & 0.137 & 0.000 & 0.159 & 0.000 & 0.159 & 0.000 \\
    21 & Accra & 0.989 & 0.000 & 0.011 & 0.000 & 0.011 & 0.000 & 0.011 & 0.000 \\
    22 & BuenosAires & 0.999 & 0.000 & 0.001 & 0.000 & 0.001 & 0.000 & 0.001 & 0.000 \\
    23 & Reykjavik & 0.932 & 0.000 & 0.068 & 0.000 & 0.073 & 0.000 & 0.073 & 0.000 \\
    24 & Bangkok & 0.997 & 0.000 & 0.003 & 0.000 & 0.003 & 0.000 & 0.003 & 0.000 \\
    25 & Vancouver & 0.944 & 0.000 & 0.056 & 0.000 & 0.060 & 0.000 & 0.060 & 0.000 \\
    26 & Almaty & 0.978 & 0.000 & 0.022 & 0.000 & 0.023 & 0.000 & 0.023 & 0.000 \\
    27 & RioDeJaneiro & 0.930 & 0.000 & 0.070 & 0.000 & 0.076 & 0.000 & 0.076 & 0.000 \\
    \bottomrule
    \end{tabular}
    \caption{Scores for the multivariate analysis of variance (MANOVA) for determining whether the distribution of available images differs significantly from the distribution of all public streets for Mapillary.}
    \label{tab:sig-mly-public}
\end{table}
\twocolumn

\end{document}